\documentclass[]{bytedance}
\usepackage[toc,page,header]{appendix}

\usepackage{minitoc}
\usepackage{amsfonts}
\usepackage{amssymb}
\usepackage{tabularx}
\usepackage{listings}
\usepackage{xcolor}
\usepackage{cancel}

\usepackage{tabulary,multirow,xspace}
\usepackage{fixmath,mathtools,nicefrac,mmstyle}
\usepackage{subcaption}
\usepackage{caption}
\usepackage{wrapfig} 
\usepackage[misc]{ifsym} 
\usepackage{colortbl}

\usepackage{wrapfig}
\usepackage{multicol}
\usepackage[most]{tcolorbox}
\usepackage{pifont}
\usepackage{tikz}
\definecolor{codegreen}{rgb}{0,0.6,0}
\definecolor{codegray}{rgb}{0.5,0.5,0.5}
\definecolor{codepurple}{rgb}{0.58,0,0.82}
\definecolor{backcolour}{rgb}{0.95,0.95,0.92}
\definecolor{boxblue}{RGB}{57,89,163}
\definecolor{boxbluebg}{RGB}{230,237,250} 

\lstdefinestyle{mystyle}{
    backgroundcolor=\color{backcolour},   
    commentstyle=\color{codegreen},
    keywordstyle=\color{magenta},
    numberstyle=\tiny\color{codegray},
    stringstyle=\color{codepurple},
    basicstyle=\ttfamily\footnotesize,
    breakatwhitespace=false,         
    breaklines=true,                 
    captionpos=b,                    
    keepspaces=true,                 
    numbers=none,                    
    numbersep=5pt,                  
    showspaces=false,                
    showstringspaces=false,
    showtabs=false,                  
    tabsize=2
}
\usepackage{pifont}

\newlength\savewidth
\newcolumntype{x}[1]{>{\centering\arraybackslash}p{#1pt}}

\newcommand{\app}{\raise.17ex\hbox{$\scriptstyle\sim$}}

\makeatletter
\DeclareRobustCommand\onedot{\futurelet\@let@token\@onedot}
\def\@onedot{\ifx\@let@token.\else.\null\fi\xspace}

\makeatother

\makeatletter

\newcommand{\Rmnum}[1]{\expandafter\@slowromancap\romannumeral #1@}
\makeatother

\usepackage{xcolor}
\usepackage{graphicx}
\usepackage{amssymb}
\usepackage{pifont}
\usepackage{floatrow}
\usepackage{amsmath} 
\usepackage{float}
\usepackage{wrapfig}
\usepackage{multirow}
\usepackage{tcolorbox}
\tcbuselibrary{breakable, skins, raster}
\usepackage{listings}
\usepackage{listings}

\definecolor{commentgreen}{rgb}{0.1, 0.4, 0.1}
\definecolor{keywordblue}{rgb}{0.1, 0.1, 0.7}
\definecolor{stringred}{rgb}{0.7, 0.1, 0.1}

\lstdefinestyle{mystyle}{
    commentstyle=\color{commentgreen},
    keywordstyle=\color{keywordblue},   
    stringstyle=\color{stringred},
    basicstyle=\ttfamily\scriptsize, 
    breaklines=true,
    keepspaces=true,
    showstringspaces=false,
    frame=none,                     
    language=Python, 
}

\newcommand{\name}{Lynx}
\title{\name{}: Towards High-Fidelity Personalized Video Generation}

\author{
\centerline{
    Shen Sang $^*$ \quad
    Tiancheng Zhi $^*$ \quad
    Tianpei Gu \quad
    Jing Liu \quad
    Linjie Luo
}
}

\affiliation[]{Intelligent Creation, ByteDance}
\contribution[*]{Equal contribution}

\abstract{
We present \textbf{Lynx}, a high-fidelity model for personalized video synthesis from a single input image. Built on an open-source Diffusion Transformer (DiT) foundation model, Lynx introduces two lightweight adapters to ensure identity fidelity. The \textit{ID-adapter} employs a Perceiver Resampler to convert ArcFace-derived facial embeddings into compact identity tokens for conditioning, while the \textit{Ref-adapter} integrates dense VAE features from a frozen reference pathway, injecting fine-grained details across all transformer layers through cross-attention. These modules collectively enable robust identity preservation while maintaining temporal coherence and visual realism. Through evaluation on a curated benchmark of 40 subjects and 20 unbiased prompts, which yielded 800 test cases, Lynx has demonstrated superior face resemblance, competitive prompt following, and strong video quality, thereby advancing the state of personalized video generation.
}

\date{\today}

\checkdata[Project Page]{\url{https://byteaigc.github.io/Lynx/}}

\begin{document}
\maketitle

\vspace{-3mm}

\begin{figure}[h]
\centering
\begin{subfigure}{0.5\textwidth}
        \centering
\begin{tikzpicture}
  \node[rounded corners=6pt, clip, inner sep=0pt]
    {\includegraphics[width=\linewidth]{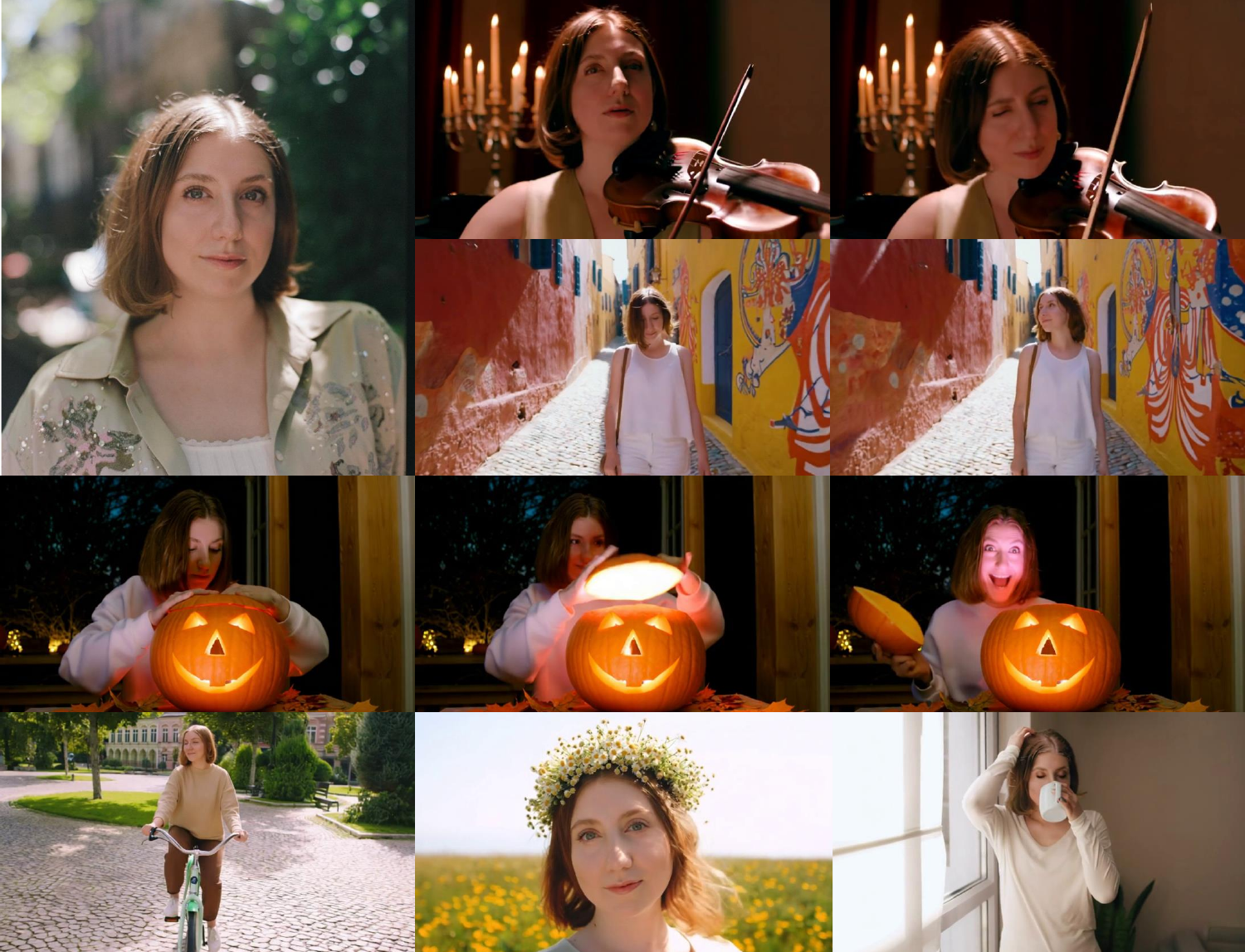}};
\end{tikzpicture}
\end{subfigure}
    \hfill
    \begin{subfigure}{0.46\textwidth}
        \centering
        \includegraphics[width=\linewidth]{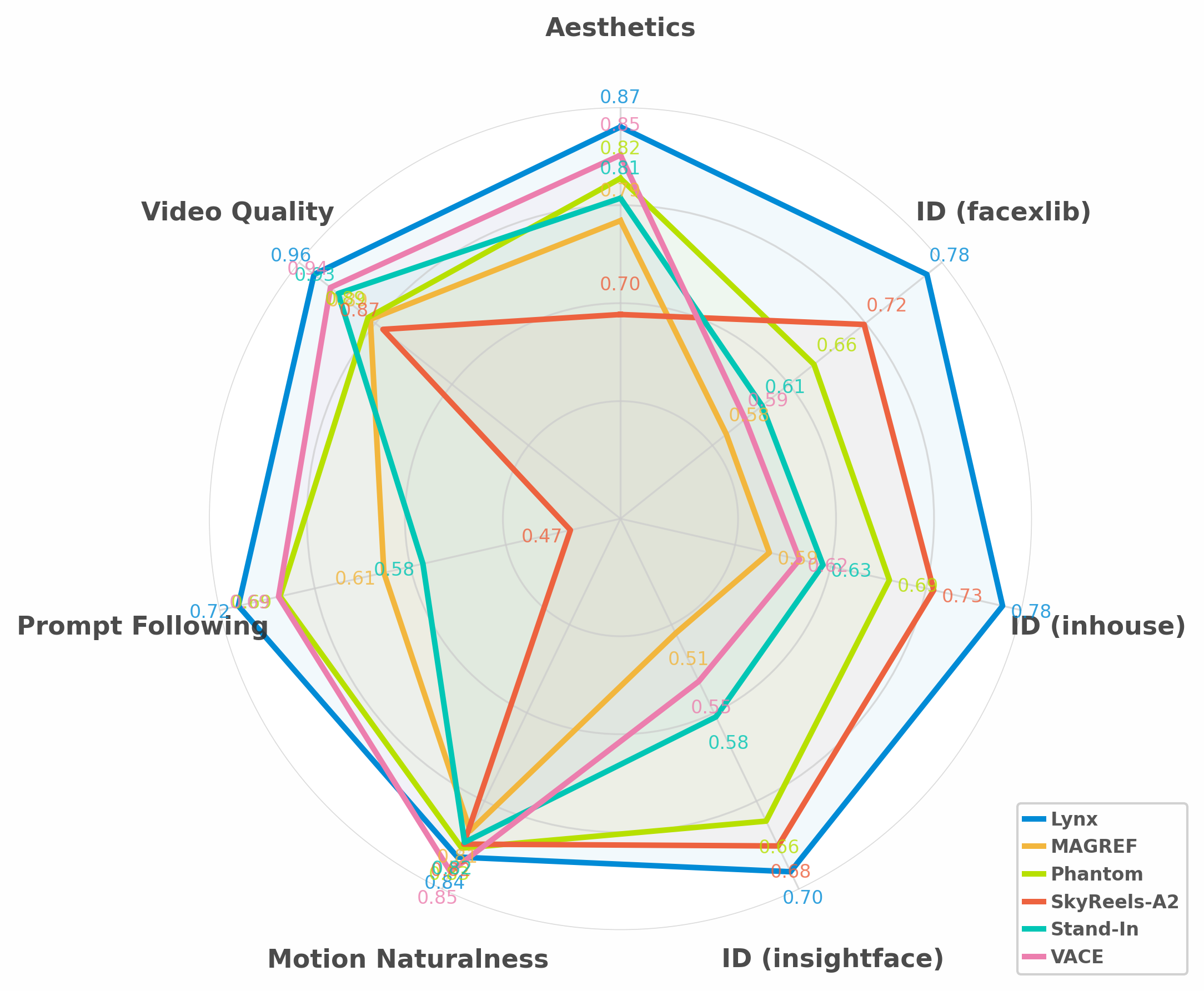}
    \end{subfigure}
    \caption{Left: Lynx consistently preserves facial identity with high fidelity, while producing natural motion, coherent lighting, and flexible scene adaptation (input shown at top-left). Right: Lynx demonstrates clear superiority in identity resemblance and perceptual quality, while remaining competitive in motion naturalness compared to other methods.}
\label{fig:radar_chart}
\end{figure}

\section{Introduction}
\label{sec:intro}

\begin{figure}[htbp]
    \centering
    \setlength{\tabcolsep}{2pt} 
    \renewcommand{\arraystretch}{0.8} 
    \begin{tabular}{cccc}
        \includegraphics[height=0.16\linewidth]{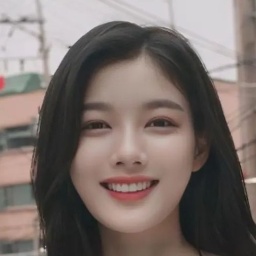} &
        \includegraphics[height=0.16\linewidth]{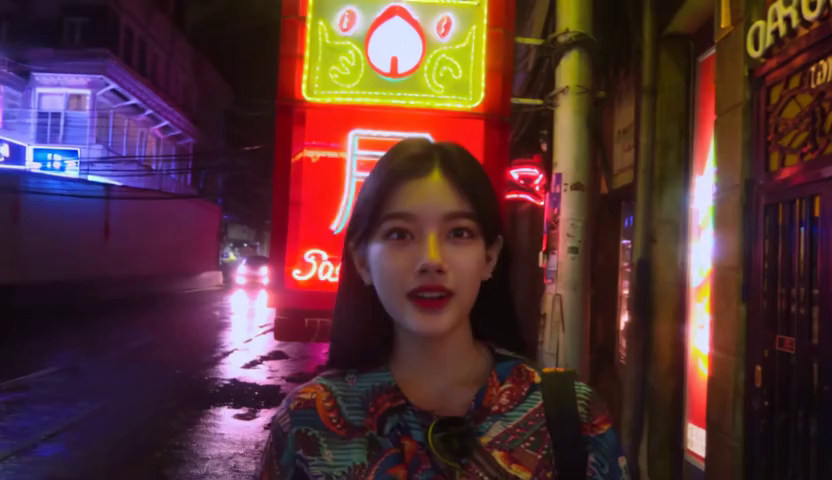} &
        \includegraphics[height=0.16\linewidth]{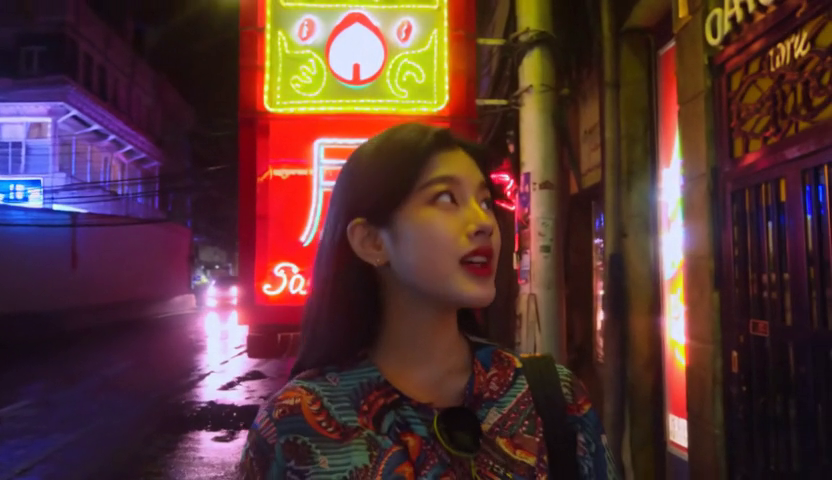} &
        \includegraphics[height=0.16\linewidth]{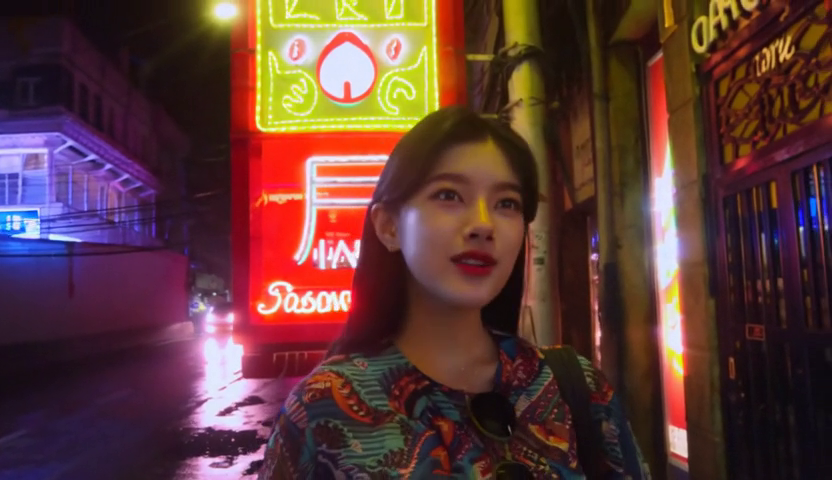} \\        \includegraphics[height=0.16\linewidth]{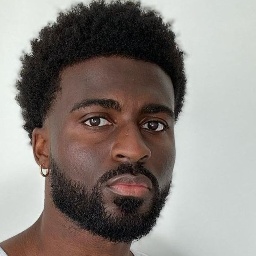} &
        \includegraphics[height=0.16\linewidth]{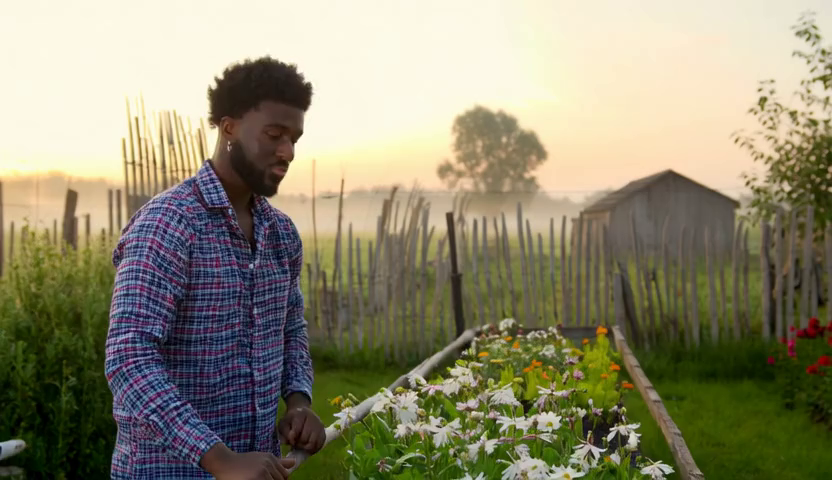} &
        \includegraphics[height=0.16\linewidth]{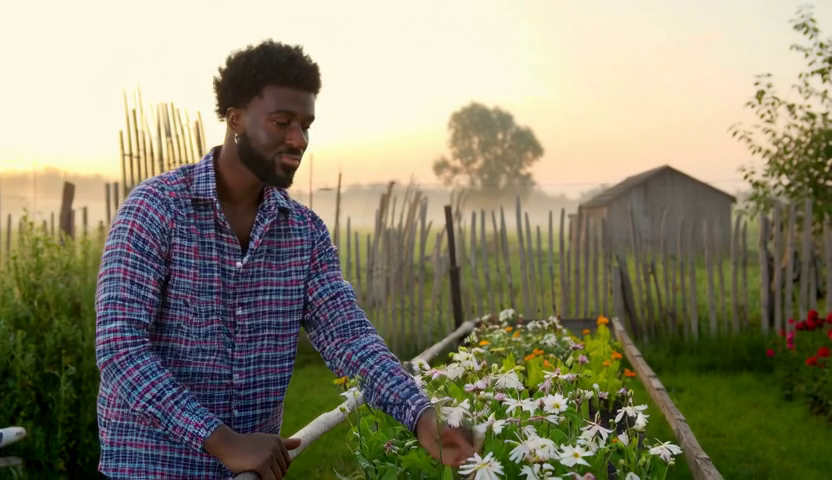} &
        \includegraphics[height=0.16\linewidth]{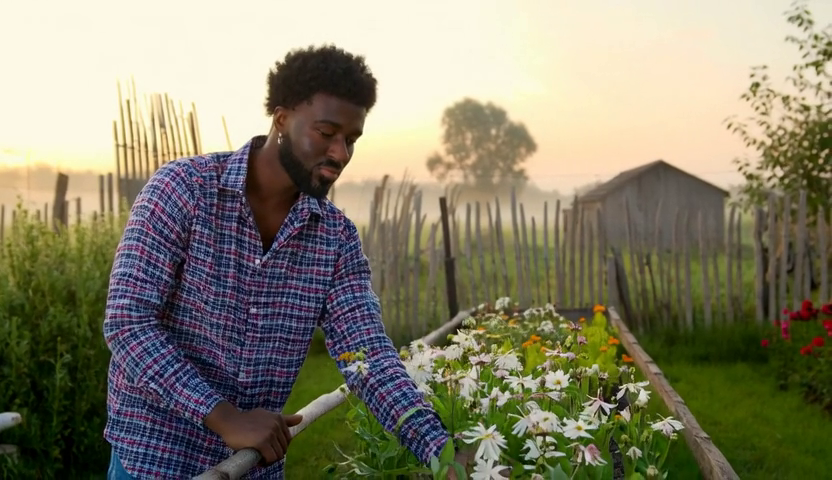} \\
        \includegraphics[height=0.16\linewidth]{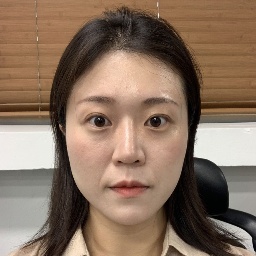} &
        \includegraphics[height=0.16\linewidth]{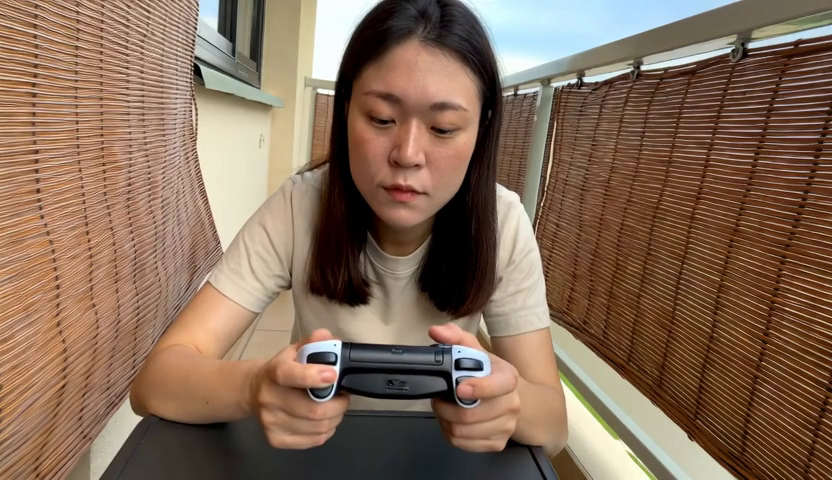} &
        \includegraphics[height=0.16\linewidth]{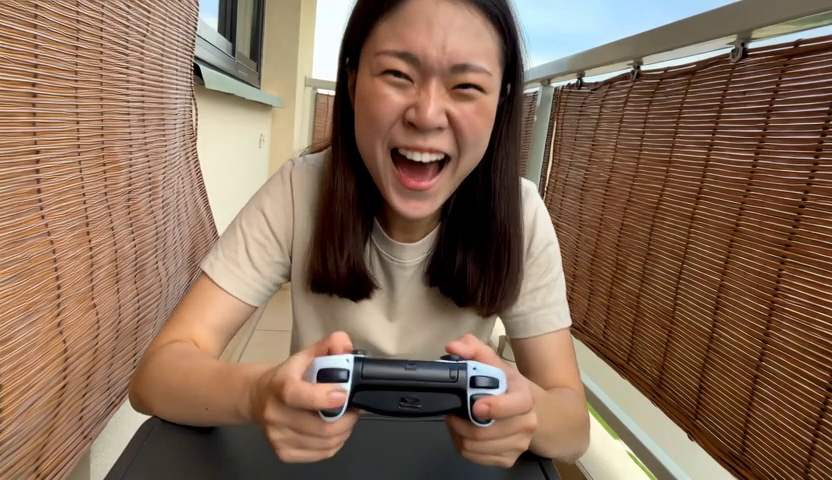} &
        \includegraphics[height=0.16\linewidth]{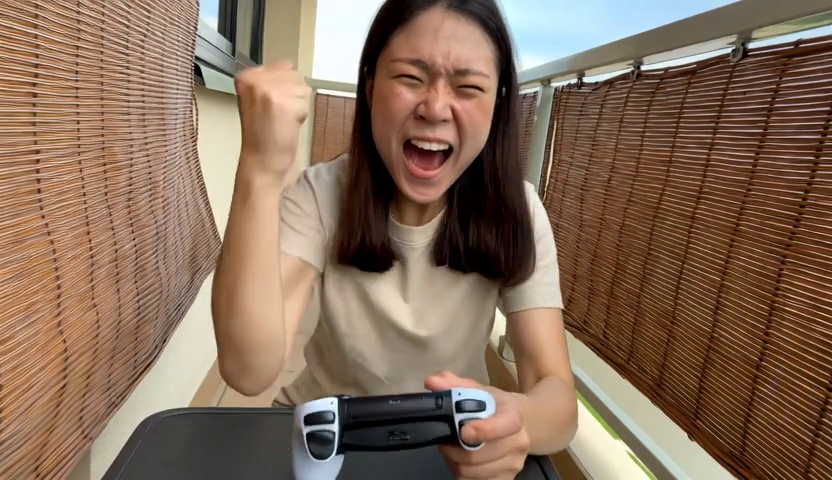} \\
        \includegraphics[height=0.16\linewidth]{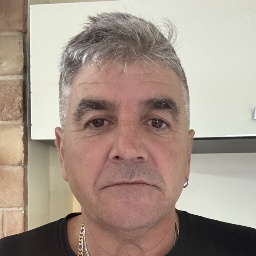} &
        \includegraphics[height=0.16\linewidth]{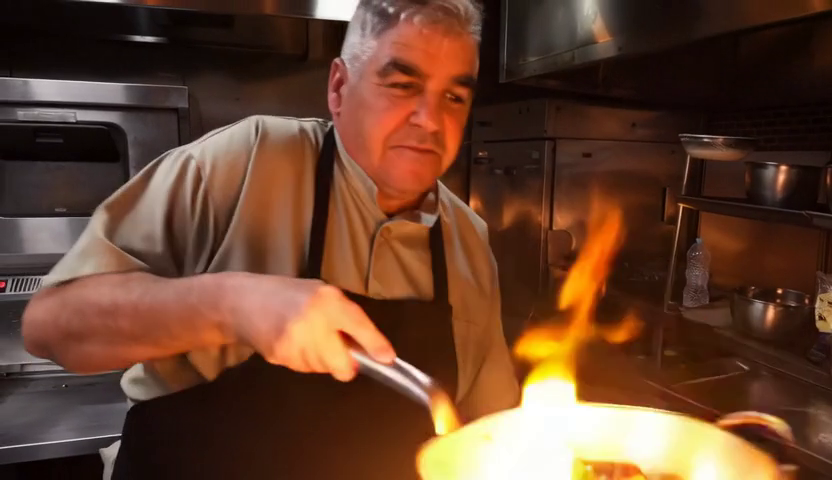} &
        \includegraphics[height=0.16\linewidth]{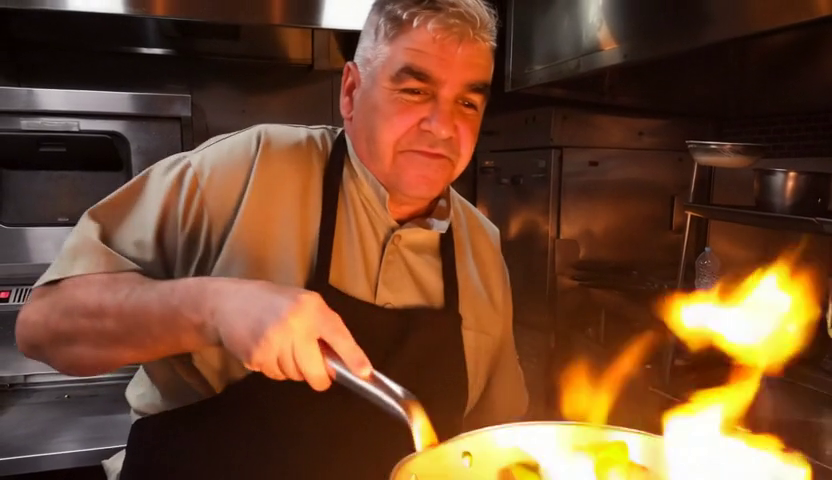} &
        \includegraphics[height=0.16\linewidth]{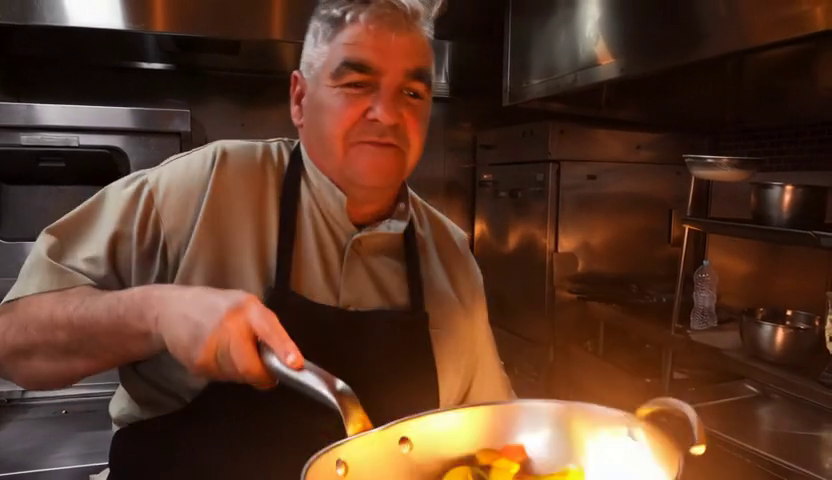} \\
        \includegraphics[height=0.16\linewidth]{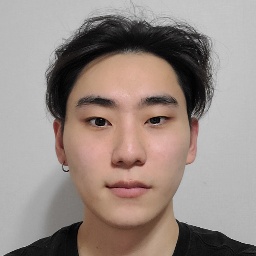} &
        \includegraphics[height=0.16\linewidth]{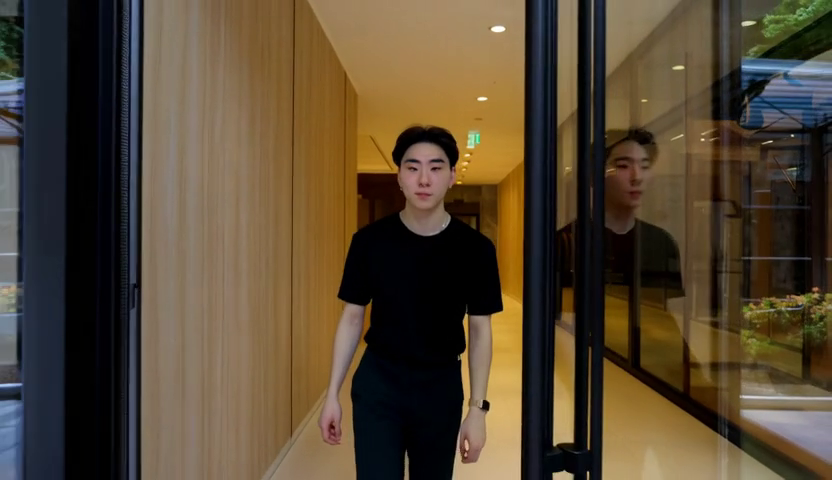} &
        \includegraphics[height=0.16\linewidth]{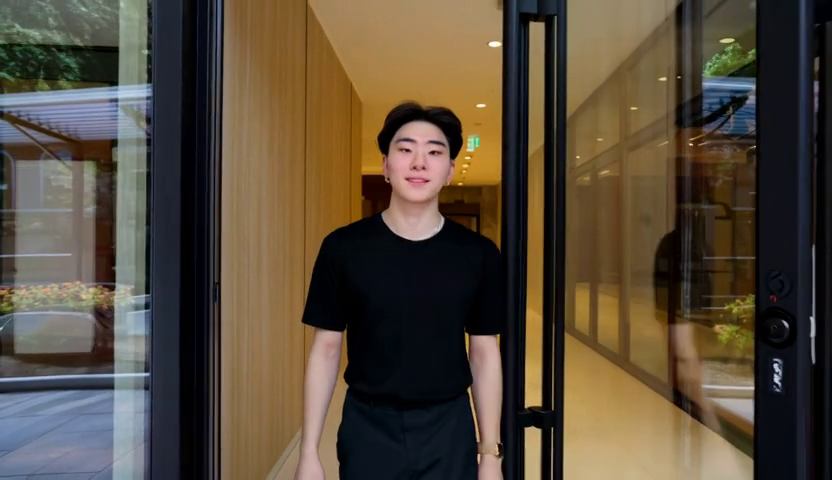} &
        \includegraphics[height=0.16\linewidth]{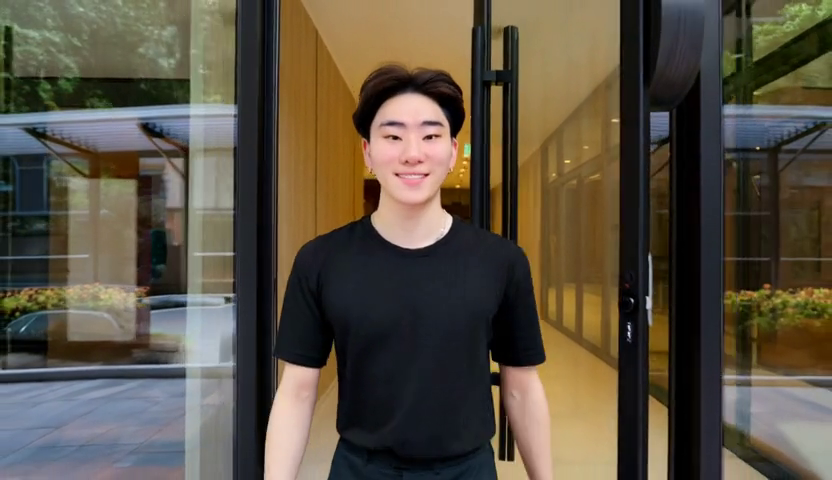} \\
        \includegraphics[height=0.16\linewidth]{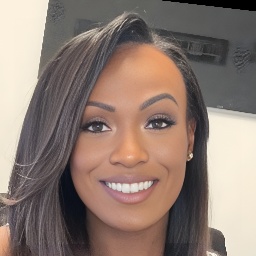} &
        \includegraphics[height=0.16\linewidth]{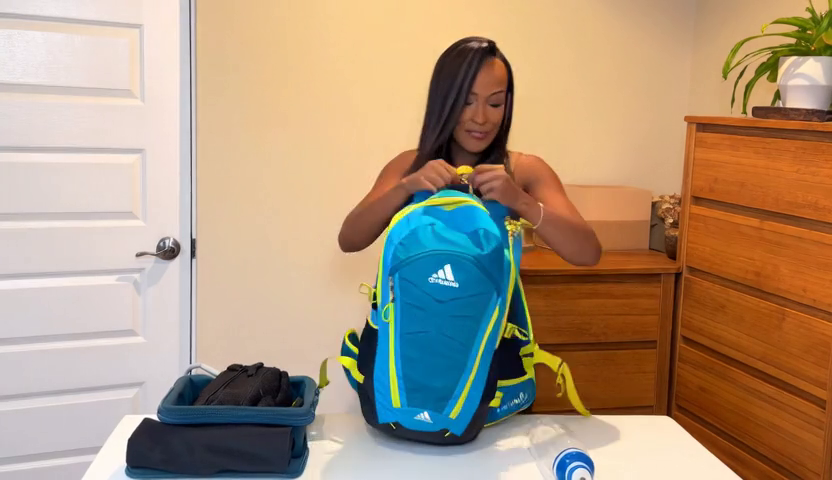} &
        \includegraphics[height=0.16\linewidth]{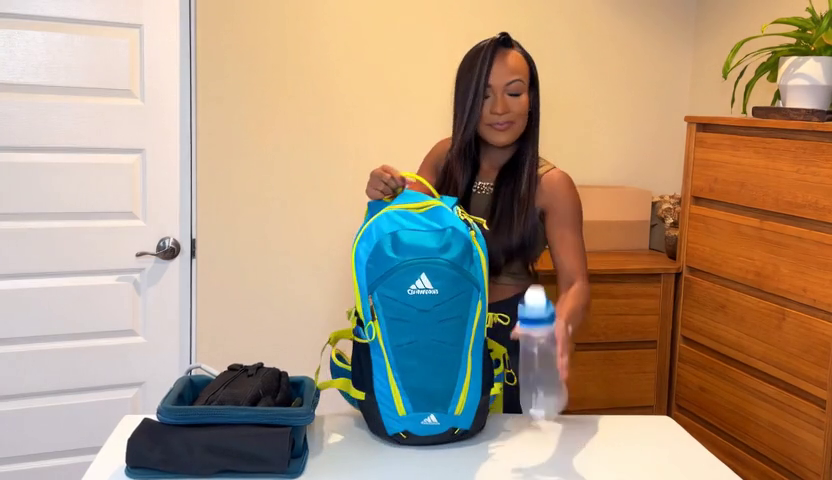} &
        \includegraphics[height=0.16\linewidth]{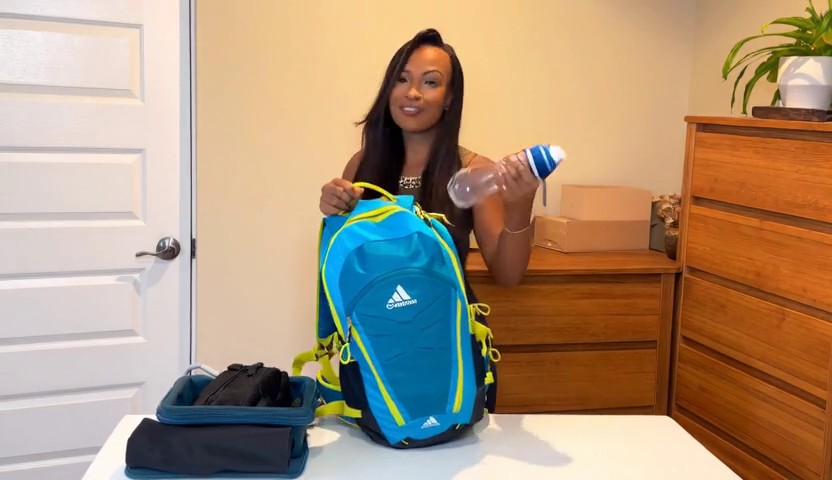} \\
        \includegraphics[height=0.16\linewidth]{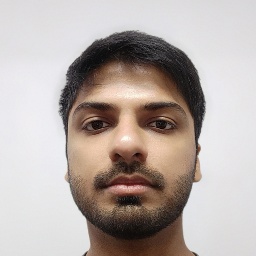} &
        \includegraphics[height=0.16\linewidth]{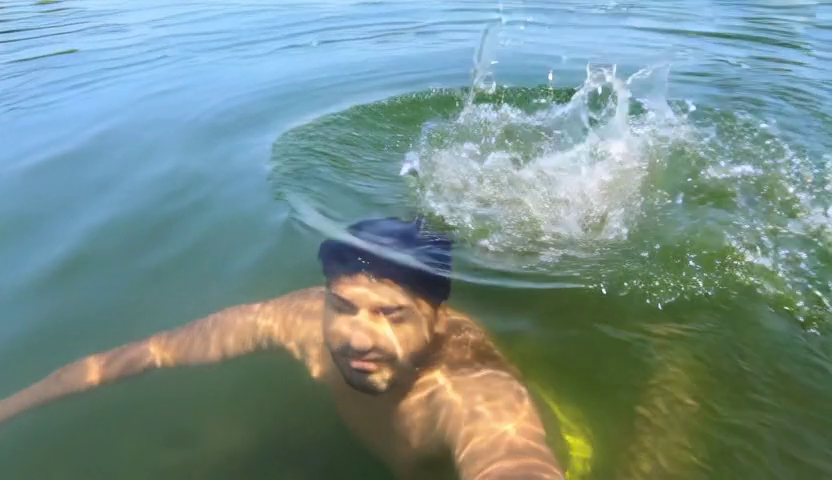} &
        \includegraphics[height=0.16\linewidth]{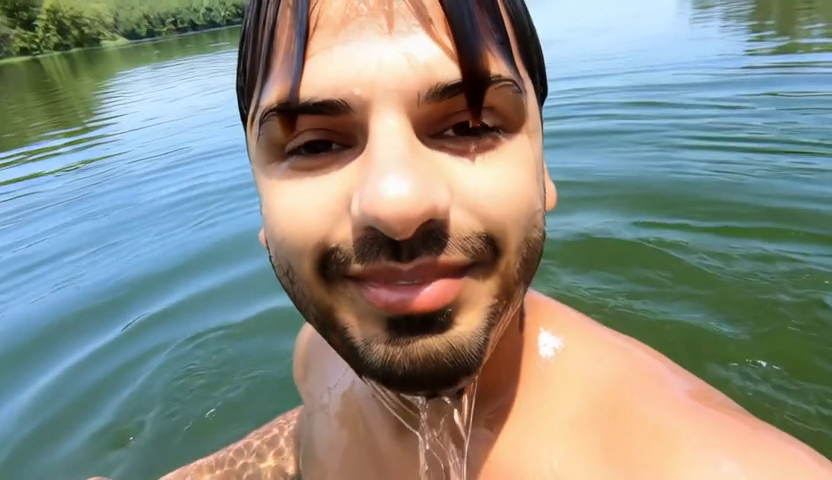} &
        \includegraphics[height=0.16\linewidth]{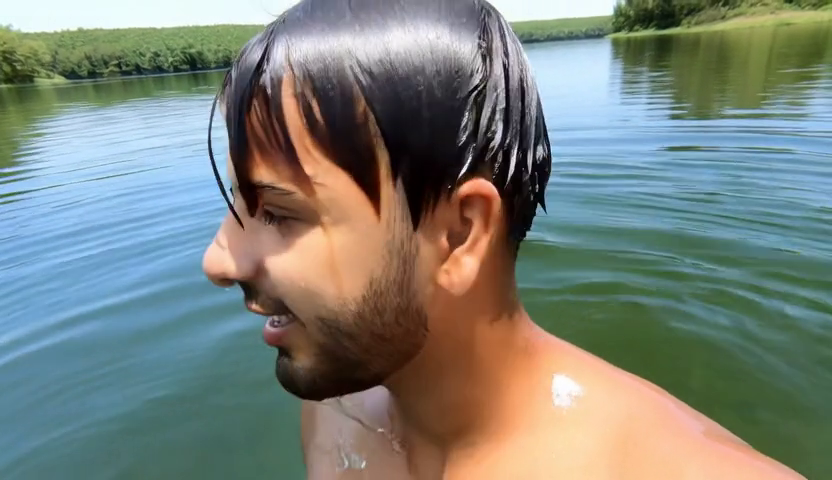} \\
        \includegraphics[height=0.16\linewidth]{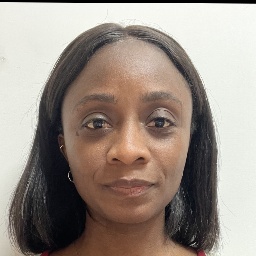} &
        \includegraphics[height=0.16\linewidth]{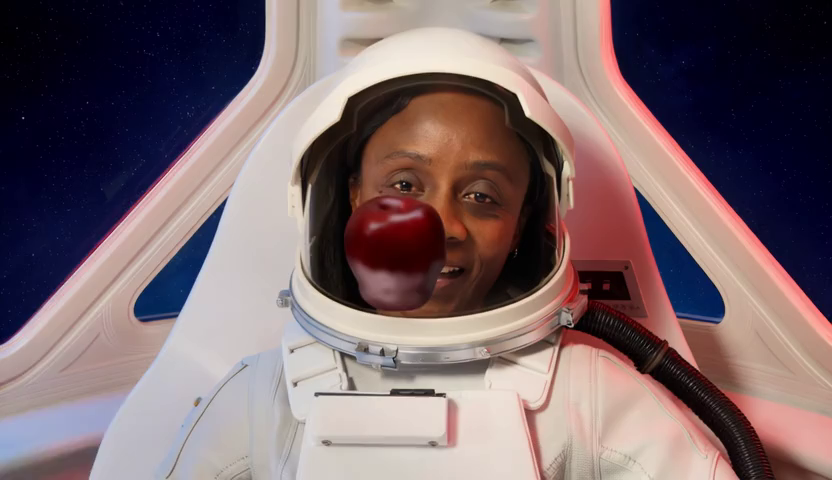} &
        \includegraphics[height=0.16\linewidth]{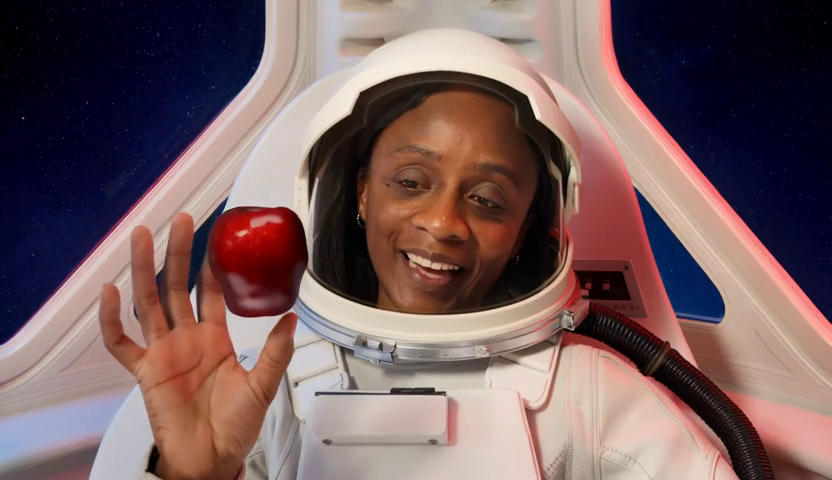} &
        \includegraphics[height=0.16\linewidth]{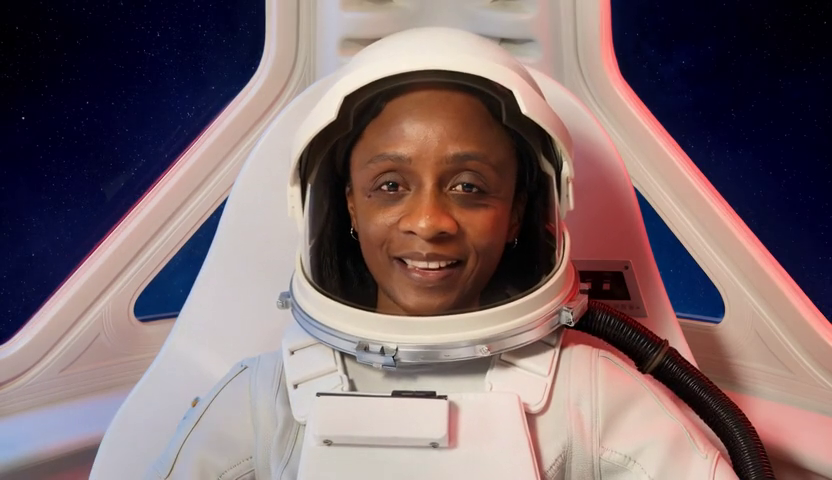} \\
    \end{tabular}
\caption{Videos generated from a single input image, showing strong identity preservation across expressive facial expressions (rows 3), diverse lighting (rows 1, 4, 5), pose variations (rows 2, 6, 7), and object interactions (rows 8).}
    \label{fig:teaser}
\end{figure}

The field of visual content generation has witnessed rapid progress, largely propelled by the emergence of diffusion models~\cite{ho2020denoising, rombach2022high, peebles2023scalable}, which offer a scalable and effective framework for high-fidelity synthesis across diverse modalities. Building upon early breakthroughs in text-to-image generation~\cite{ramesh2021zero, nichol2021glide, rombach2021highresolution, saharia2022photorealistic, batifol2025flux}, the community has extended diffusion-based methods into the temporal domain, giving rise to text-to-video models~\cite{singer2022make, blattmann2023stable, kondratyuk2023videopoet, sora2024, yang2024cogvideox, kong2024hunyuanvideo, wan2025, veo2025} capable of synthesizing dynamic visual content from natural language prompts. Recent advancements in backbone architectures—such as Diffusion Transformers (DiT)~\cite{peebles2023scalable}—have further improved generation quality and scalability. Beyond foundational generation, there is growing interest in downstream tasks including video editing~\cite{li2025flowdirector0, zhang2024effived, wang2025videodirector}, multi-shot storytelling~\cite{kara2025shotadapter}, and controllable motion synthesis~\cite{jin2025flovd, geng2025motion}, reflecting the field’s increasing demand for controllability, reusability, and efficiency. A key direction emerging from this trend is personalized video generation, which aims to synthesize videos that faithfully preserve subject identity.

Personalized content creation has been extensively explored in the image domain, where foundation generative models are adapted to user-provided references to enable the synthesis of diverse yet identity-consistent outputs. Early methods~\cite{ruiz2023dreambooth, gal2022image, hu2022lora} achieve strong identity preservation via either full-model or parameter-efficient fine-tuning such as LoRA~\cite{hu2022lora}. More recent approaches~\cite{ye2023ip-adapter, wang2024instantid, zhang2025idpatch} employ lightweight conditioning mechanisms based on identity embeddings or reference features, enabling efficient personalization without retraining the entire model. Building upon these advances, as well as the progress in general-purpose video foundation models, recent efforts have begun to explore identity preservation in the temporal domain~\cite{yuan2025identity, fei2025skyreels, liu2025phantom, vace, deng2025magref, xue2025standin}, aiming to generate coherent and realistic personalized videos over time.

In this report, we present \textbf{Lynx}, a high-fidelity personalized video generation framework designed to preserve identity from a single input image. Instead of restructuring or fine-tuning the full base model, Lynx adopts an adapter-based design with two specialized components: the \textit{ID-adapter} and the \textit{Ref-adapter}. The \textit{ID-adapter} leverages cross-attention to inject identity features extracted from a single facial image. Specifically, facial embeddings are obtained using a face recognition model and transformed into a compact set of identity tokens via a Perceiver Resampler, enabling rich and efficient representation learning. To further enhance detail preservation, the \textit{Ref-adapter} incorporates reference features extracted from a pretrained VAE encoder (inherited from the base model). These features are passed through a frozen copy of the diffusion backbone to obtain intermediate activations across all DiT blocks, which are then fused into the generation process via cross-attention. For training, we adopt a multi-stage progressive strategy with a spatio-temporal frame packing design to effectively handle both image and video data of varying aspect ratios and temporal lengths.

We evaluate Lynx on a curated benchmark of 40 diverse subjects and 20 human-centric unbiased prompts, resulting in 800 test cases. Face resemblance is assessed using three expert face recognition models. To evaluate prompt following and video quality, we construct an automated pipeline with the Gemini-2.5-Pro API~\footnote{https://ai.google.dev/gemini-api/docs}, instructing the model to score aesthetic quality, motion naturalness, prompt alignment, and overall video quality. As reported in Table~\ref{tab:quant_comp} and Table~\ref{tab:quant_comp_gemini}, Lynx consistently outperforms recent state-of-the-art personalized video generation methods in identity preservation, while also delivering stronger prompt alignment and superior overall video quality.
\section{Related Works}
\label{sec:related}

\textbf{Video Foundation Models.}
Recent video foundation models are predominantly built on the diffusion framework, where variational autoencoders (VAEs)~\cite{kingma2013auto} compress raw videos into compact latent representations, enabling efficient training and generation. Early latent diffusion methods extended image foundation models with U-Net architectures to the video domain by incorporating temporal modules such as 3D convolutions and temporal attention~\cite{singer2022make, blattmann2023stable, guo2023animatediff}. As the demand for scalability and long-range temporal coherence increased, research shifted toward transformer-based architectures. Diffusion Transformers (DiT)~\cite{peebles2023scalable} and their dual-stream variant MMDiT~\cite{esser2024scaling} demonstrated more expressive spatio-temporal modeling, leading to improved temporal consistency. These architectures now underpin state-of-the-art video foundation models, including CogVideoX~\cite{yang2024cogvideox}, HunyuanVideo~\cite{kong2024hunyuanvideo}, Wan2.1~\cite{wan2025}, Seedance~1.0~\cite{gao2025seedance}, \textit{etc.}, which achieve strong generalization through large-scale training data, substantial computational resources, and extended context length.

\textbf{Identity-Preserving Content Creation.}
Identity-preserving generation is a central topic in content creation and has been extensively studied in the image domain. Early approaches~\cite{ruiz2023dreambooth, gal2022image, hu2022lora} typically rely on model fine-tuning or optimization to obtain subject-specific models. However, such tuning-based methods are often impractical for real-world applications because of their computational cost and lack of scalability. For example, DreamBooth~\cite{ruiz2023dreambooth} and LoRA-based variants~\cite{hu2022lora} require fine-tuning either the full base model or additional low-rank adapters. To overcome these limitations, tuning-free methods~\cite{ye2023ip-adapter, wang2024instantid} introduce lightweight ID-injection modules that avoid per-subject training. IP-Adapter~\cite{ye2023ip-adapter} represents identity features with a face recognition encoder and injects them into the base model through adapters. Building on this idea, InstantID~\cite{wang2024instantid} incorporates a ControlNet~\cite{zhang2023adding} module for input decoupling and finer-grained control.

With the advent of large video foundation models, research attention has shifted toward personalized video generation. For instance, ConsistID~\cite{yuan2025identity} enforces facial identity consistency via frequency decomposition. ConceptMaster~\cite{huang2025conceptmaster} employs a CLIP image encoder and a learnable Q-Former to fuse visual representations with corresponding text embeddings for each concept. HunyuanCustom~\cite{hu2025hunyuancustom} extends HunyuanVideo~\cite{kong2024hunyuanvideo} with a multi-modal customization framework that integrates image, audio, video, and text conditions through modality-specific modules, achieving stronger identity consistency and controllable video generation. Another line of work (\textit{e.g.}, SkyReels-A2~\cite{fei2025skyreels}, VACE~\cite{vace}, Phantom~\cite{liu2025phantom}) concatenates reference conditions with noisy latents and processes the full sequence during denoising. However, balancing identity resemblance and editability has long been a persistent challenge. Our method significantly improves resemblance while maintaining strong prompt following and high video quality.
\section{Architecture and Training Strategy}
\label{sec:method}

\subsection{Model Architecture}
\begin{figure}[t]
    \centering
    \includegraphics[width=1.0\linewidth]{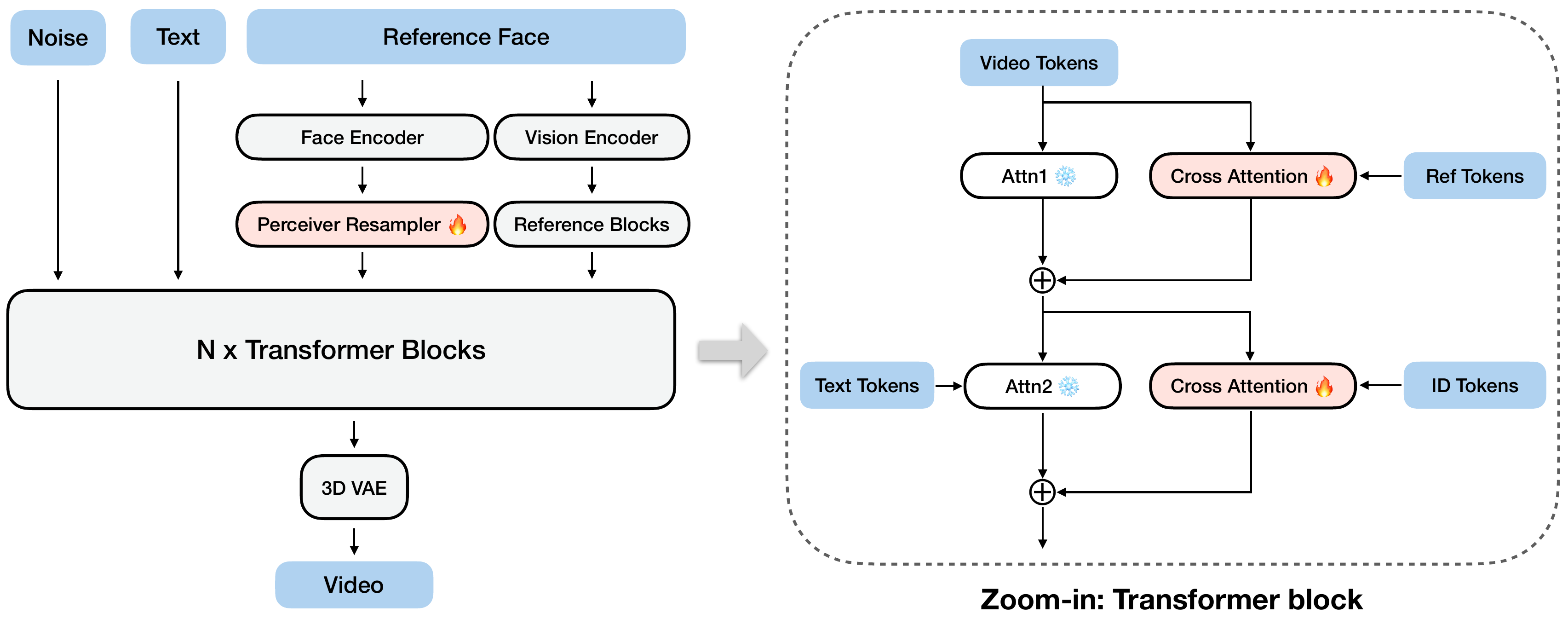}
    \caption{Architecture of Lynx. Built on a DiT-based video foundation model, Lynx introduces two adapter modules that inject identity features through cross-attention.}
    \label{fig:arch}
\end{figure}

We adopt Wan2.1~\cite{wan2025}, one of the latest open-sourced video foundation models, as our base model. Wan is built upon the DiT architecture~\cite{peebles2023scalable}, combined with the Flow Matching~\cite{lipman2022flow} framework. Each DiT block first applies spatio-temporal self-attention over visual tokens, enabling joint modeling of spatial details and temporal dynamics, followed by cross-attention to incorporate text conditions.

Instead of restructuring and fine-tuning the full model, we introduce two adapter modules, \textit{i.e.}, \textit{ID-adapter} and \textit{Ref-adapter}, to inject identity features and enable personalized video generation on top of the base model. The overall architecture and adapter design are illustrated in Figure~\ref{fig:arch}.

\textbf{ID-adapter.} Prior works~\cite{ye2023ip-adapter, wang2024instantid} incorporated face recognition features~\cite{deng2019arcface} to achieve personalized generation in text-to-image models such as Stable Diffusion~\cite{rombach2021highresolution}. These methods typically attach additional adapter layers and introduce extra cross-attention modules to condition generation on identity features. Specifically, face image is passed through a face feature extractor to obtain a feature vector. To convert this vector into a sequence suitable for cross-attention, a Perceiver Resampler~\cite{alayrac2022flamingo} (also known as the Q-Former~\cite{li2023blip}) is trained to map it into a fixed-length token embedding representation. We adopt the same paradigm. Given a face feature vector of dimension 512, the Resampler produces a sequence of 16 token embeddings of dimension 5120. The token embedding is concatenated with 16 additional register tokens~\cite{darcet2023vision} and cross-attended with the input visual tokens. The resulting representation is then added back to the main branch.

\textbf{Ref-adapter.} Several recent approaches~\cite{fei2025skyreels, liu2025phantom} use VAE features to enhance detail preservation during reference injection, taking advantage of the spatially dense representations produced by VAE encoders. Complementing the ID-adapter, our design also incorporates VAE dense features to enhance identity fidelity. Unlike prior approaches that directly place the feature map in front of noisy latents in an image-to-image-like generation fashion, we instead process the reference image through a frozen copy of the base model (with noise level as 0 and text prompt as "image of a face"), similar to the design of ReferenceNet~\cite{hu2023animateanyone}. This allows spatial details from the reference image to be captured across all layers. As with ID-adapter, we apply separate cross-attention at each layer to integrate the corresponding reference tokens.

\subsection{Training Strategy}

We describe here the strategies employed for large-scale training. Since training videos (and images) vary in both spatial resolution and temporal duration, we adopt the NaViT approach~\cite{dehghani2023patch} to efficiently batch heterogeneous inputs. Multiple videos or images are packed into a single long sequence, with attention masks applied to separate samples. Training follows a progressive curriculum beginning with image pretraining, which leverages the abundance of large-scale image data, and is then extended to video training to restore temporal dynamics.

\subsubsection{Spatio-Temporal Frame Pack}

Traditional training in the image domain often relies on bucketing to handle multi-resolution inputs. Images are cropped and resized into a set of predefined aspect ratios and resolutions, and during training the data loader samples from a single bucket so that images within a batch share the same dimensions. While effective for images, this strategy does not generalize well to video, as the additional temporal dimension (frame length) introduces significant complexity. Bucketing by both resolution and duration reduces flexibility and limits the model’s ability to generalize to arbitrary aspect ratios and video lengths.  

To overcome this limitation, inspired by Patch n’ Pack~\cite{dehghani2023patch}, we concatenate the patchified tokens of each video into a single long sequence, treating the collection a unified batch. An attention mask ensures that tokens only attend within their own video, preventing cross-sample interference. For positional encoding, we apply 3D Rotary Position Embeddings (3D-RoPE)~\cite{su2104enhanced} independently to each video. This design enables efficient batching of heterogeneous images and videos while preserving both spatial and temporal consistency.

\subsubsection{Progressive Training}

\textbf{Image Pretraining.} We begin with image pretraining given the large amount of available image data. To ensure consistency across training stages, each image is treated as a single-frame video and the same frame pack strategy described above is applied. In our experiments, training the Perceiver Resampler from scratch yielded unsatisfactory results: no facial resemblance was observed even after substantial training, suggesting that the model either fails to converge or requires prohibitively longer training. Instead, we found that initializing the Resampler from an image-domain pretrained checkpoint (\textit{e.g.}, InstantID~\cite{wang2024instantid}) leads to much faster convergence. With this initialization, recognizable facial resemblance emerges after only 10k iterations, while the complete first stage runs for 40k iterations.

\textbf{Video Training.} Image pretraining alone tends to produce videos that are largely static, as the model primarily learns to preserve appearance rather than capture motion. To restore temporal dynamics, a second stage that exposes the model to large-scale video data is necessary. This stage enables the network to learn motion patterns, scene transitions, and temporal consistency while retaining and enhancing the strong identity conditioning established during image pretraining. Training proceeds for 60k iterations. 
\section{Data Pipeline}
\label{sec:dataset}

\begin{figure}[t]
    \centering
    \begin{subfigure}[t]{0.45\linewidth}
        \centering
        \includegraphics[width=0.45\linewidth]{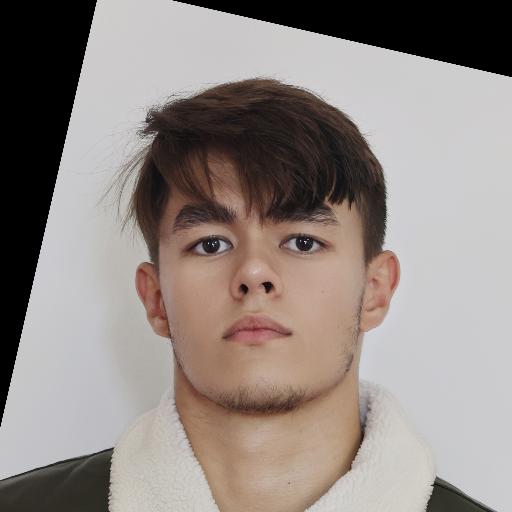}
        \includegraphics[width=0.45\linewidth]{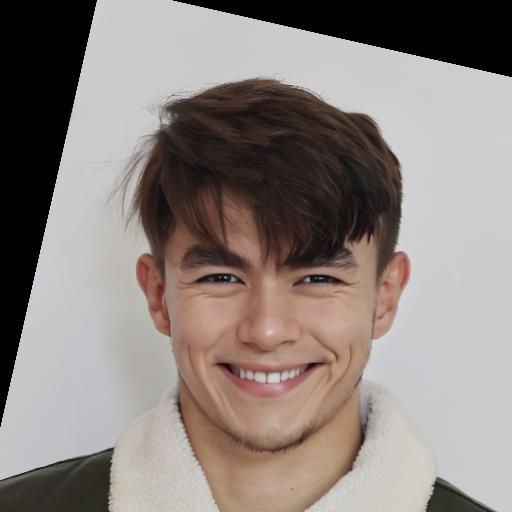}
        \caption{Expression Augmentation}
        \label{fig:xnemo}
    \end{subfigure}
    \hfill
    \begin{subfigure}[t]{0.45\linewidth}
        \centering
        \includegraphics[width=0.45\linewidth]{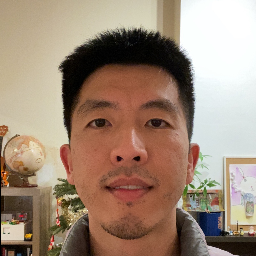}
        \includegraphics[width=0.45\linewidth]{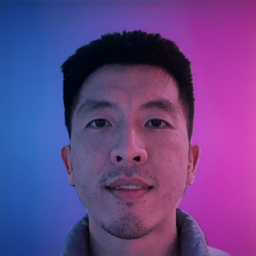}
        \caption{Portrait Relighting}
        \label{fig:relight}
    \end{subfigure}
    \caption{Examples of our augmentation strategies: (a) expression augmentation via X-Nemo~\cite{zhao2025x}, and (b) portrait relighting via LBM~\cite{chadebec2025lbm}.}
    \label{fig:augmentation}
\end{figure}

The goal of our data pipeline is to construct high-quality person–text–video triplets. While text prompts can be readily obtained through captioning models (e.g., Qwen 2.5-VL~\cite{bai2025qwen2}), the main challenge lies in establishing reliable person–video pairs, i.e., pairing an image of a person as the identity (ID) condition with a target video of the same individual.

Our raw data consist of images and videos collected from both publicly available datasets and in-house sources. These data can be categorized into four types: (1) single images; (2) single videos; (3) multi-scene image collections of the same person; and (4) multi-scene video collections of the same person. To construct image–image and image–video pairs—where one image serves as the ID condition and the other image or video serves as the generation target—a straightforward approach is to crop faces directly from images or videos. However, this often leads to overfitting of expression and lighting. Meanwhile, multi-scene data, which are essential for robust training, are inherently scarce.

To address these limitations, we adopt two augmentation strategies, illustrated in Figure~\ref{fig:augmentation}:  

\begin{itemize}
    \item \textbf{Expression Augmentation.}  
    We employ X-Nemo~\cite{zhao2025x} to edit a source face so that it matches the target expression, thereby enriching expression diversity (Figure~\ref{fig:augmentation}a).  

    \item \textbf{Portrait Relighting.}  
    We apply LBM~\cite{chadebec2025lbm} to relight faces and replace backgrounds under varying illumination conditions, enhancing robustness to lighting variation (Figure~\ref{fig:augmentation}b).  
\end{itemize}


After augmentation, we perform identity verification using a face recognition model and discard pairs with low resemblance to ensure high-quality ID consistency. Resemblance filter is also applied to raw multi-scene data without augmentation.

Finally, our pipeline constructs a total of 50.2M pairs, consisting of 21.5M single-scene pairs, 7.7M multi-scene pairs, and 21.0M augmented single-scene pairs. For single-scene pairs where the condition image is directly cropped from the target, we additionally apply background augmentation by segmenting the human subject and replacing the background. During training, these different types of pairs are retrieved through weighted sampling to balance data diversity.
\section{Experiments}
\label{sec:exp}

\begin{figure}[h]
\centering
\setlength{\tabcolsep}{0pt}

\resizebox{1.0\textwidth}{!}{%
\begin{tabular}{@{} m{0.06\textwidth} m{0.43\textwidth} m{0.012\textwidth} m{0.43\textwidth} @{}}

\mbox{} &
\begin{minipage}[t]{0.43\textwidth}
  \noindent
  \begin{minipage}[t]{0.20\linewidth}\vspace{0pt}
    \includegraphics[width=\linewidth]{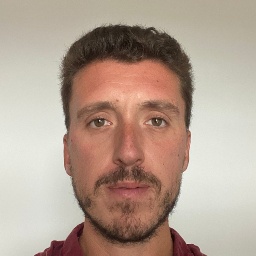}
  \end{minipage}\hspace{4pt}%
  \begin{minipage}[t]{0.77\linewidth}\vspace{3pt}\raggedright\footnotesize
    ``In a bustling open-air market filled with vibrant colors and aromatic spices, an animated person of mixed cultural background engages enthusiastically...''
  \end{minipage}
\end{minipage}
&
&
\begin{minipage}[t]{0.43\textwidth}
  \noindent
  \begin{minipage}[t]{0.2\linewidth}\vspace{0pt}
    \includegraphics[width=\linewidth]{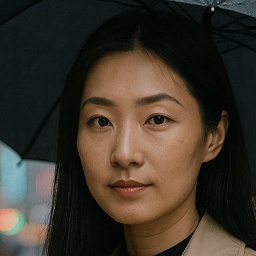}
  \end{minipage}\hspace{4pt}%
  \begin{minipage}[t]{0.77\linewidth}\vspace{3pt}\raggedright\footnotesize
    ``A person sits at a wooden table in a warmly lit kitchen, joyfully eating a plate of steaming dumplings... lift each dumpling with chopsticks...''
  \end{minipage}
\end{minipage}
\\[-2pt]

\parbox[c][0.12\textwidth][c]{0.06\textwidth}{\centering\rotatebox{90}{\scriptsize\bfseries SkyReels-A2}}
&
\begin{minipage}[c]{0.43\textwidth}
  \noindent 
  \includegraphics[width=0.32\linewidth]{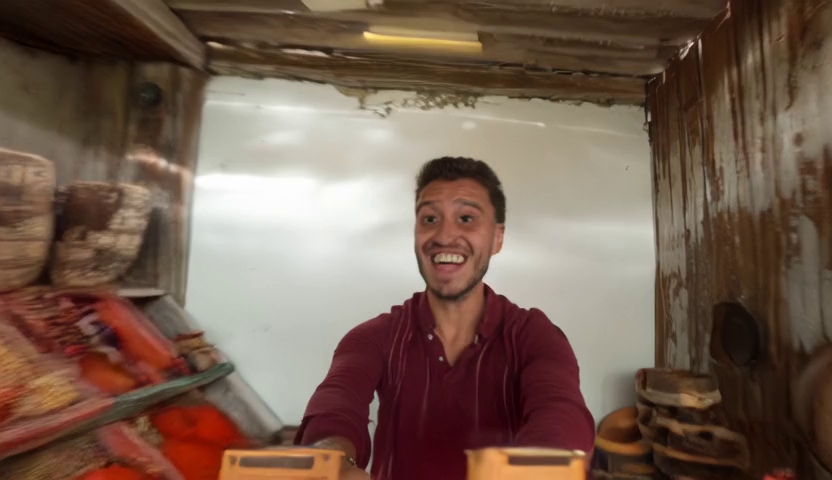}\hspace{2pt}%
  \includegraphics[width=0.32\linewidth]{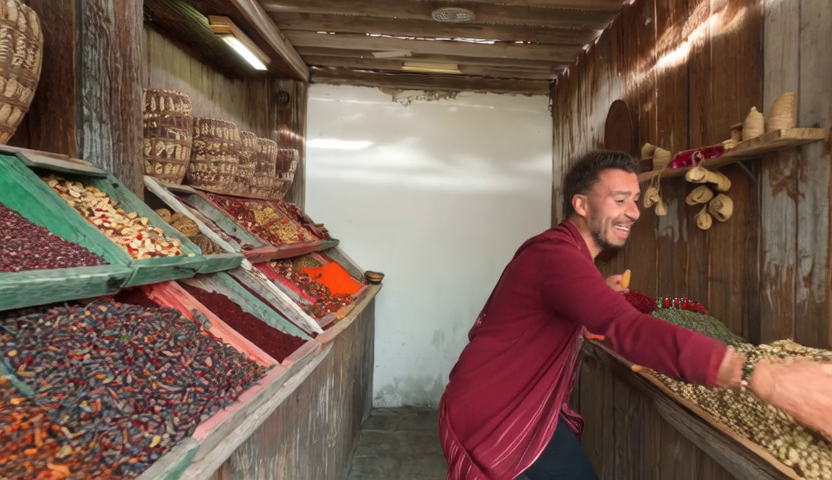}\hspace{2pt}%
  \includegraphics[width=0.32\linewidth]{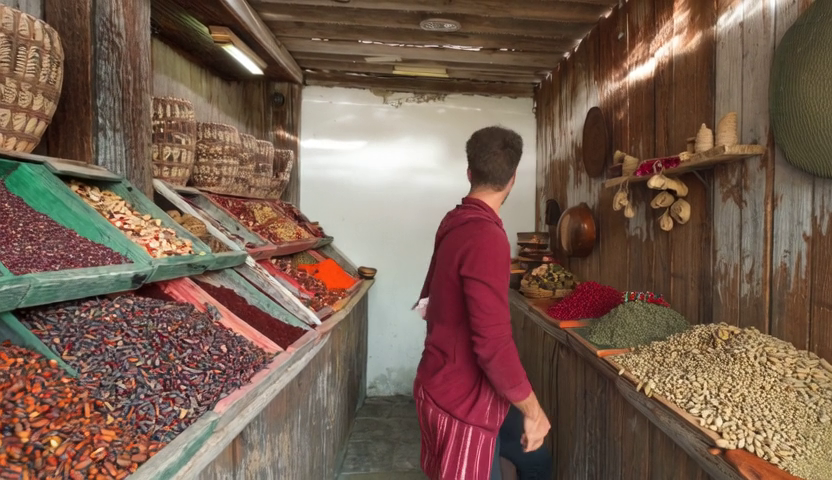}
\end{minipage}
&
&
\begin{minipage}[c]{0.43\textwidth}
  \noindent 
  \includegraphics[width=0.32\linewidth]{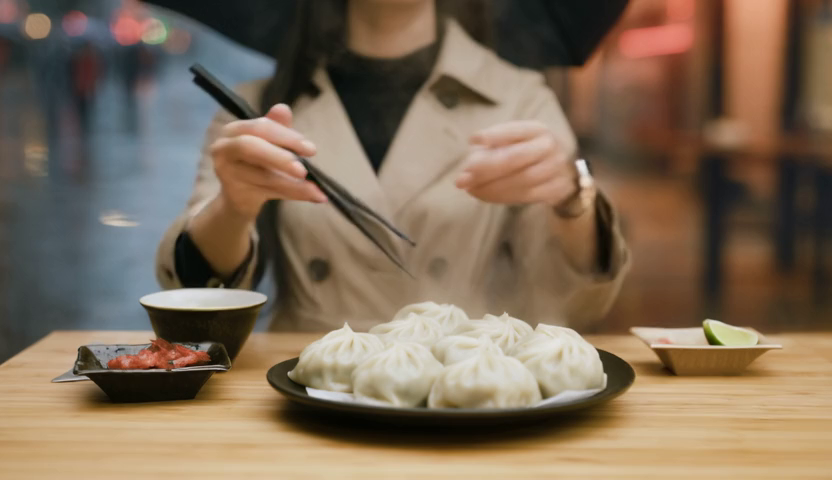}\hspace{2pt}%
  \includegraphics[width=0.32\linewidth]{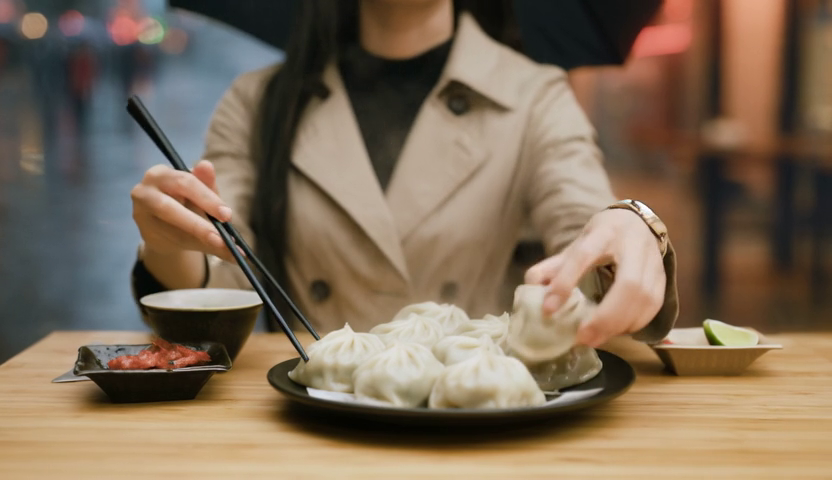}\hspace{2pt}%
  \includegraphics[width=0.32\linewidth]{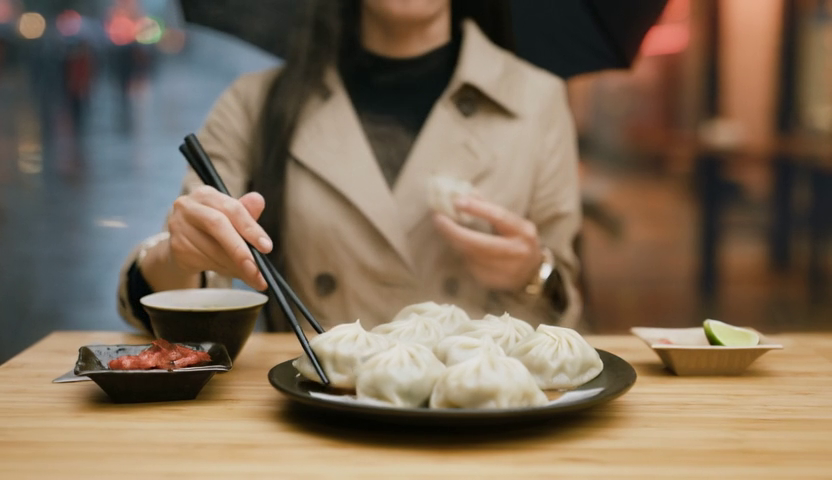}
\end{minipage}
\\[-8pt]

\parbox[c][0.12\textwidth][c]{0.06\textwidth}{\centering\rotatebox{90}{\scriptsize\bfseries VACE}}
&
\begin{minipage}[c]{0.43\textwidth}
  \noindent 
  \includegraphics[width=0.32\linewidth]{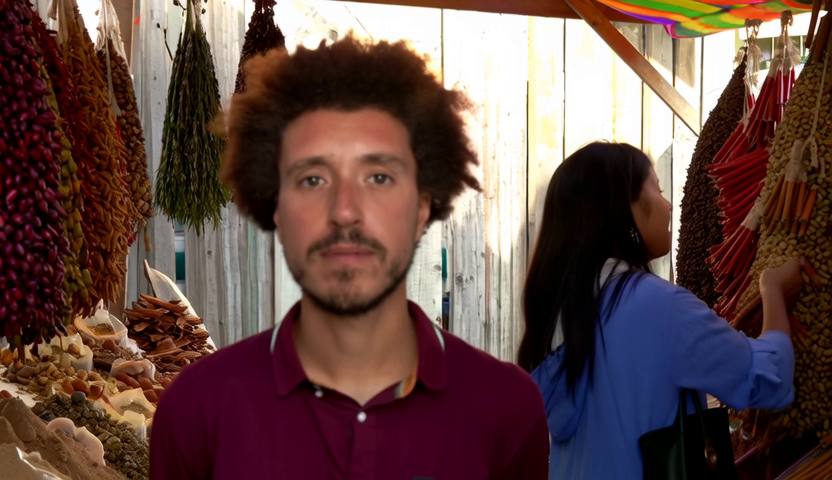}\hspace{2pt}%
  \includegraphics[width=0.32\linewidth]{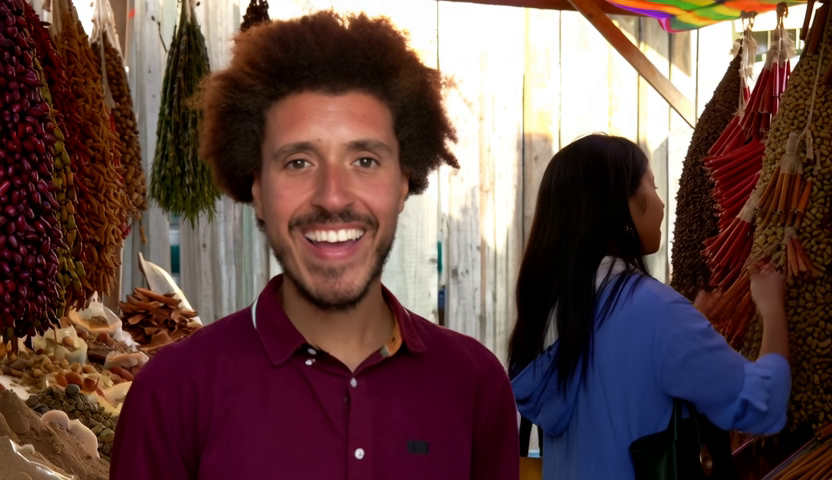}\hspace{2pt}%
  \includegraphics[width=0.32\linewidth]{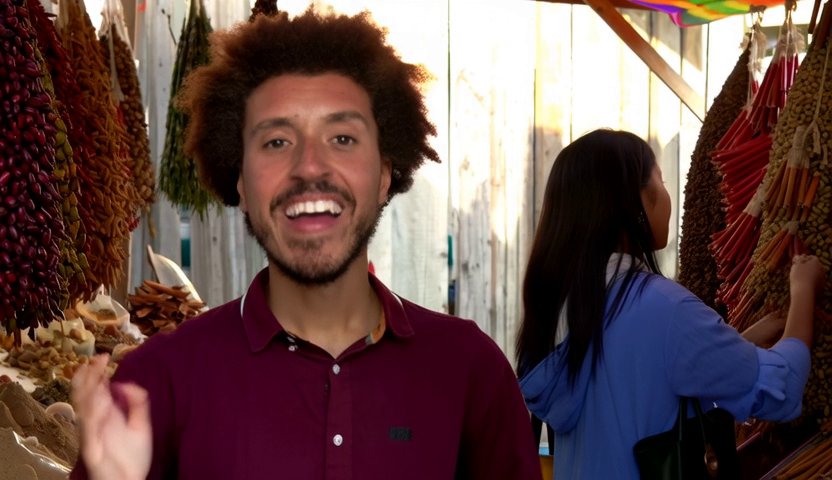}
\end{minipage}
&
&
\begin{minipage}[c]{0.43\textwidth}
  \noindent 
  \includegraphics[width=0.32\linewidth]{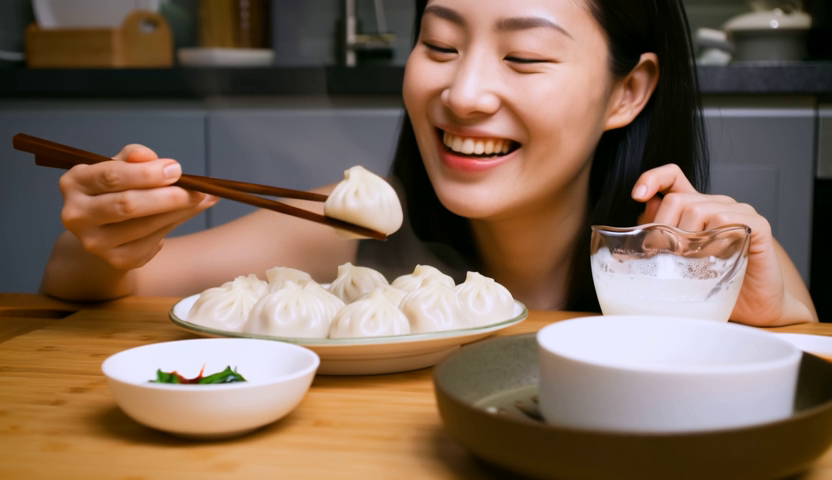}\hspace{2pt}%
  \includegraphics[width=0.32\linewidth]{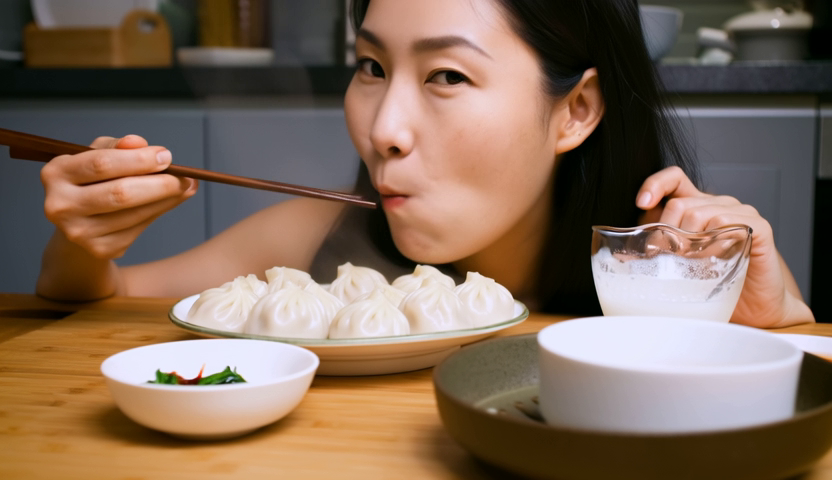}\hspace{2pt}%
  \includegraphics[width=0.32\linewidth]{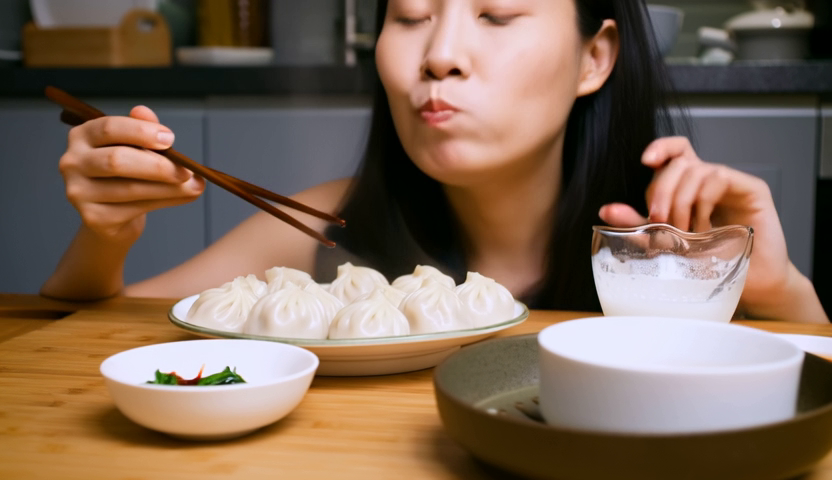}
\end{minipage}
\\[-8pt]

\parbox[c][0.12\textwidth][c]{0.06\textwidth}{\centering\rotatebox{90}{\scriptsize\bfseries Phantom}}
&
\begin{minipage}[c]{0.43\textwidth}
  \noindent 
  \includegraphics[width=0.32\linewidth]{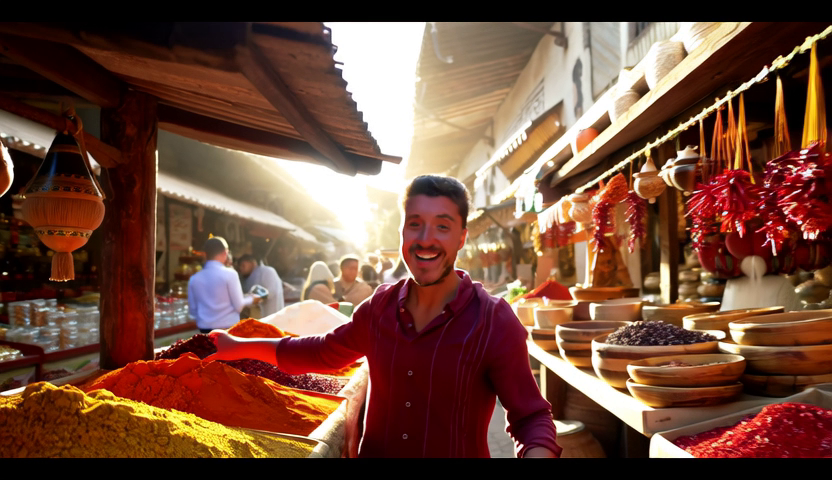}\hspace{2pt}%
  \includegraphics[width=0.32\linewidth]{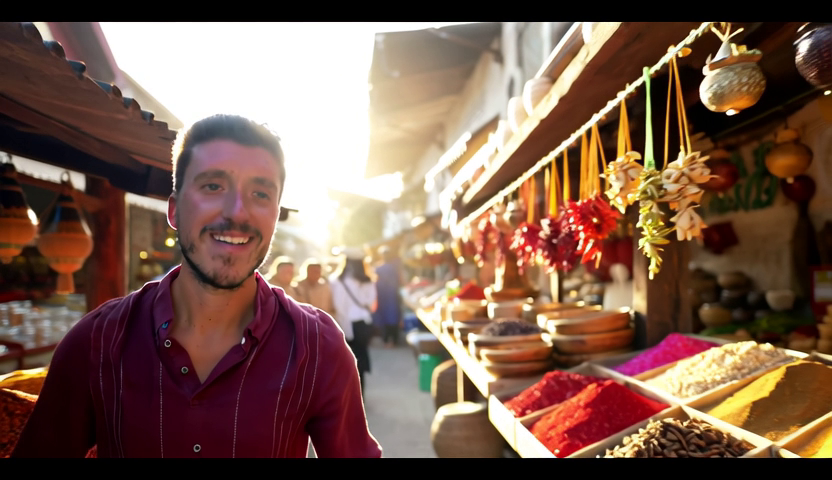}\hspace{2pt}%
  \includegraphics[width=0.32\linewidth]{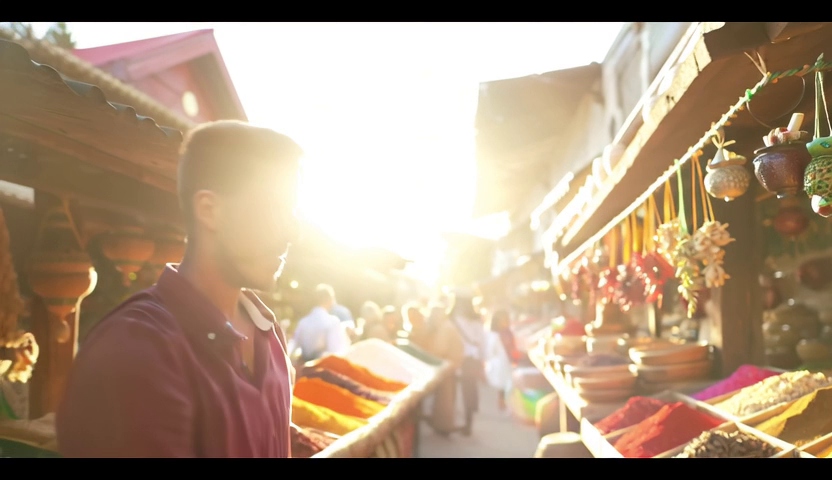}
\end{minipage}
&
&
\begin{minipage}[c]{0.43\textwidth}
  \noindent 
  \includegraphics[width=0.32\linewidth]{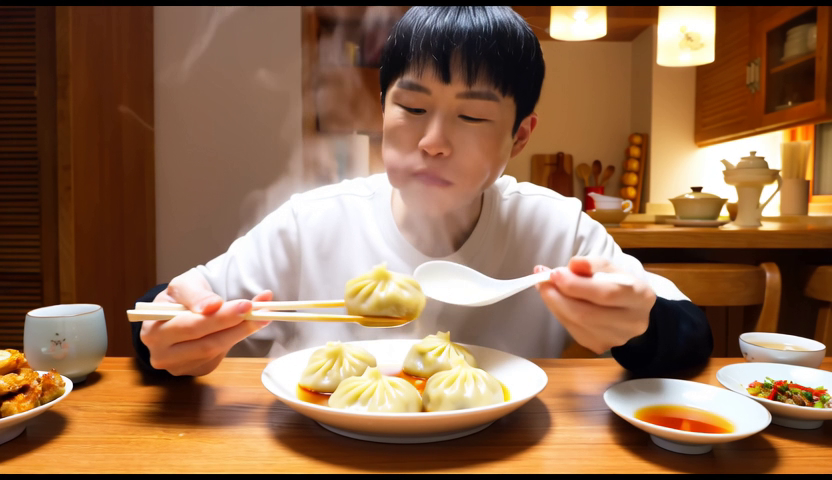}\hspace{2pt}%
  \includegraphics[width=0.32\linewidth]{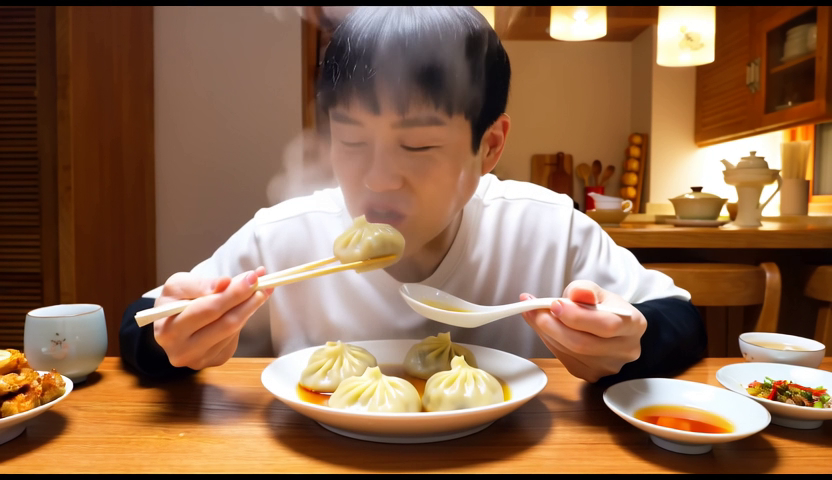}\hspace{2pt}%
  \includegraphics[width=0.32\linewidth]{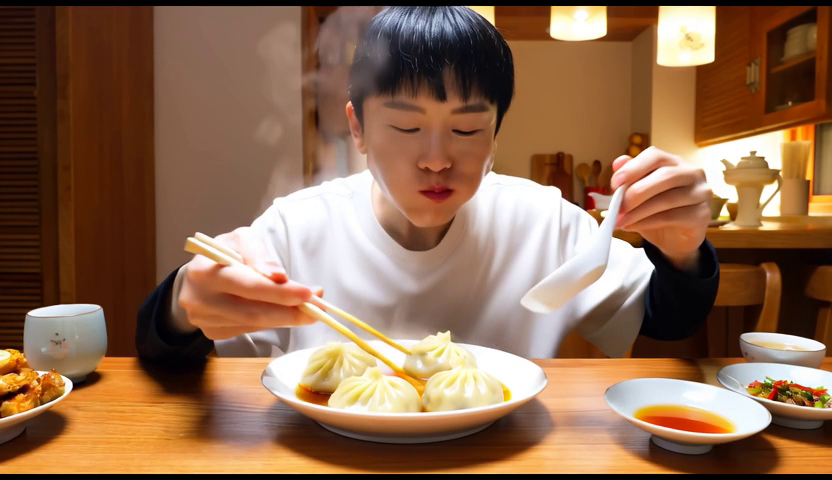}
\end{minipage}
\\[-8pt]

\parbox[c][0.12\textwidth][c]{0.06\textwidth}{\centering\rotatebox{90}{\scriptsize\bfseries MAGREF}}
&
\begin{minipage}[c]{0.43\textwidth}
  \noindent 
  \includegraphics[width=0.32\linewidth]{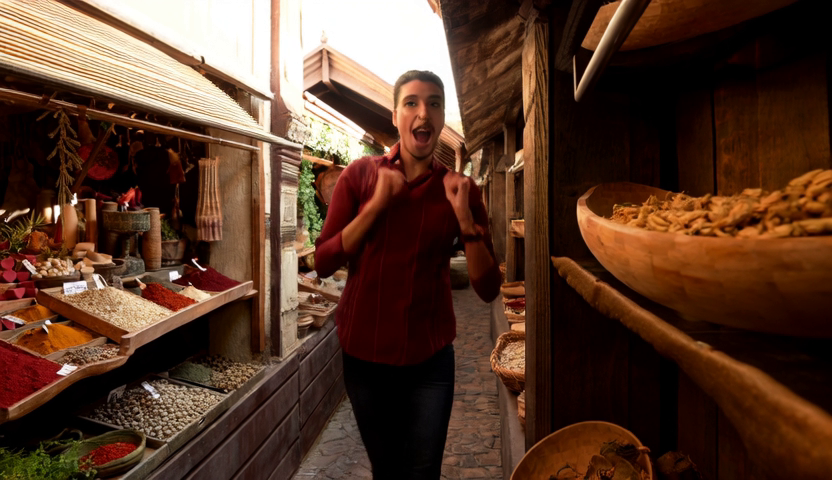}\hspace{2pt}%
  \includegraphics[width=0.32\linewidth]{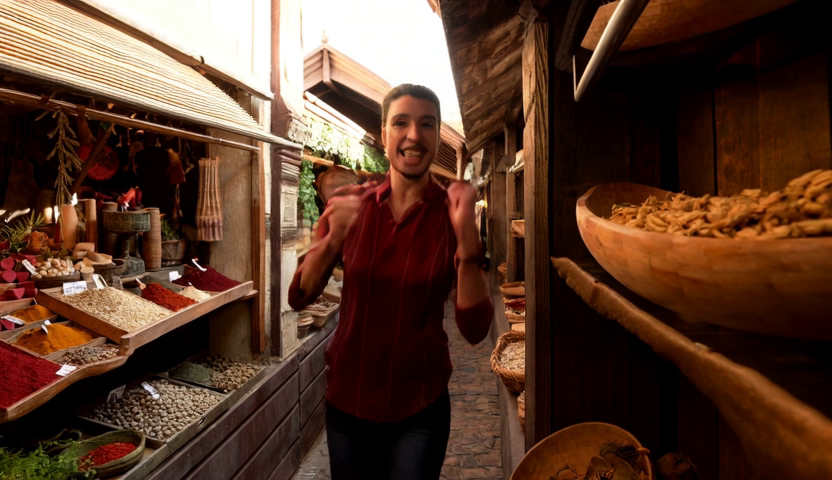}\hspace{2pt}%
  \includegraphics[width=0.32\linewidth]{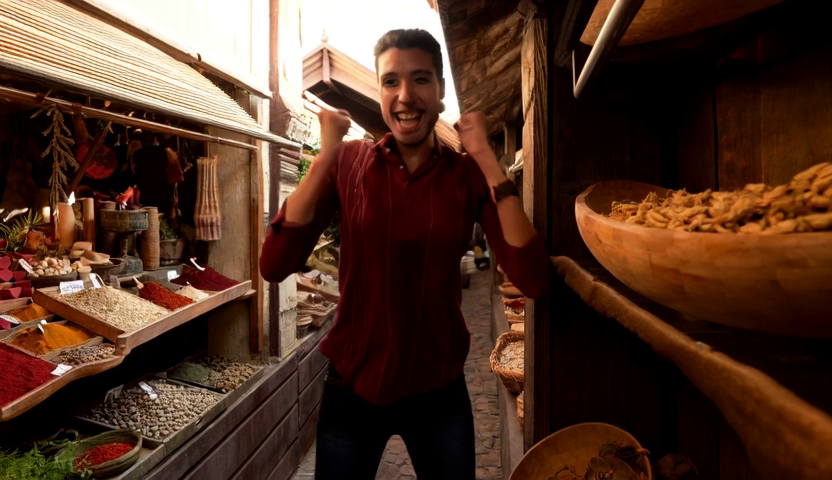}
\end{minipage}
&
&
\begin{minipage}[c]{0.43\textwidth}
  \noindent 
  \includegraphics[width=0.32\linewidth]{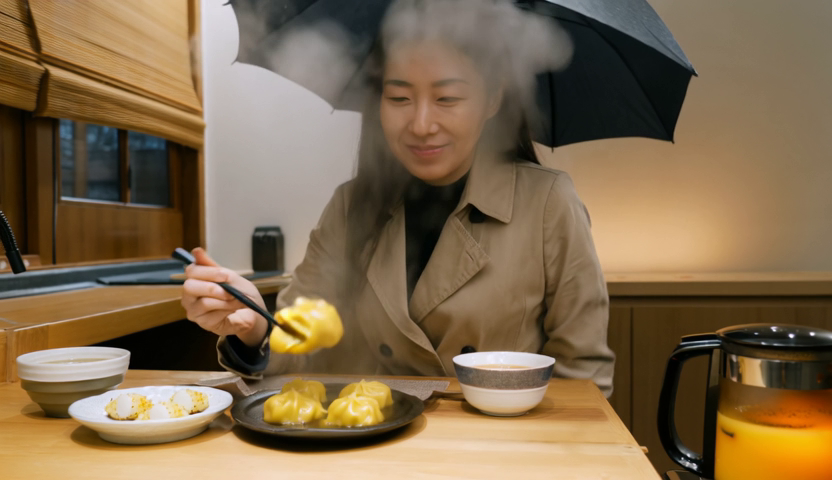}\hspace{2pt}%
  \includegraphics[width=0.32\linewidth]{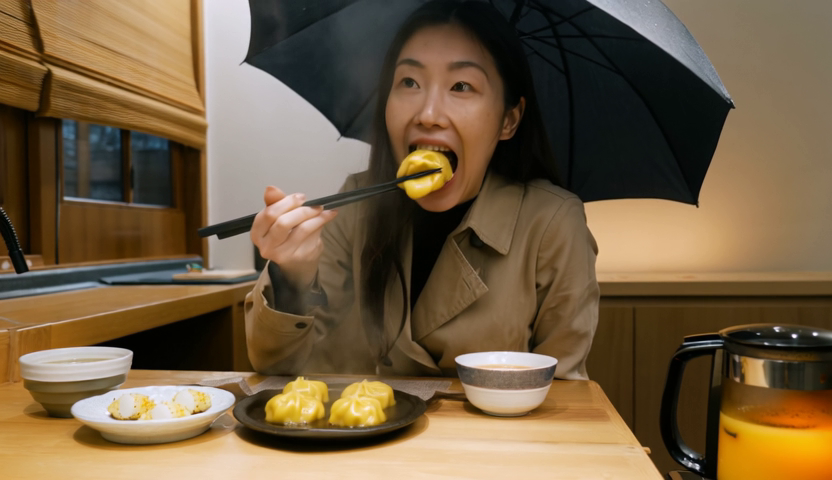}\hspace{2pt}%
  \includegraphics[width=0.32\linewidth]{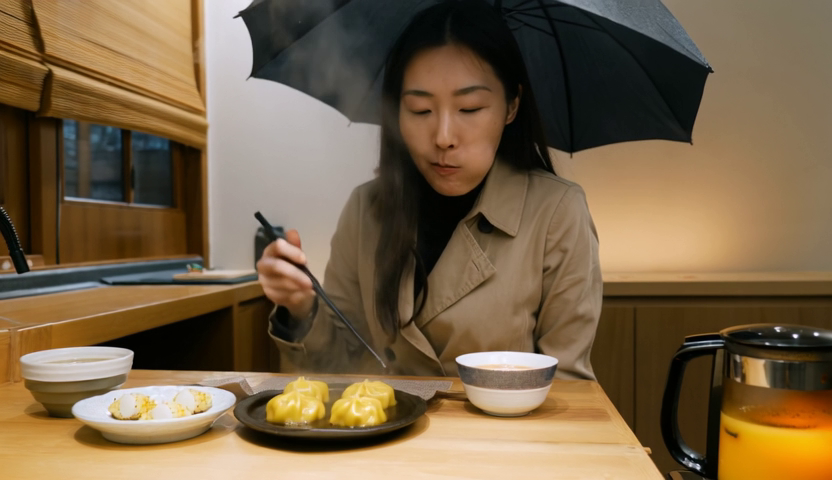}
\end{minipage}
\\[-8pt]

\parbox[c][0.12\textwidth][c]{0.06\textwidth}{\centering\rotatebox{90}{\scriptsize\bfseries Stand-In}}
&
\begin{minipage}[c]{0.43\textwidth}
  \noindent 
  \includegraphics[width=0.32\linewidth]{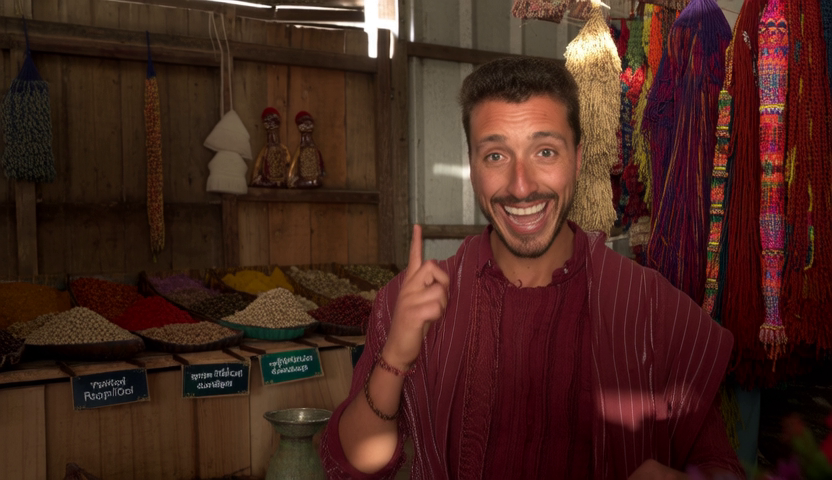}\hspace{2pt}%
  \includegraphics[width=0.32\linewidth]{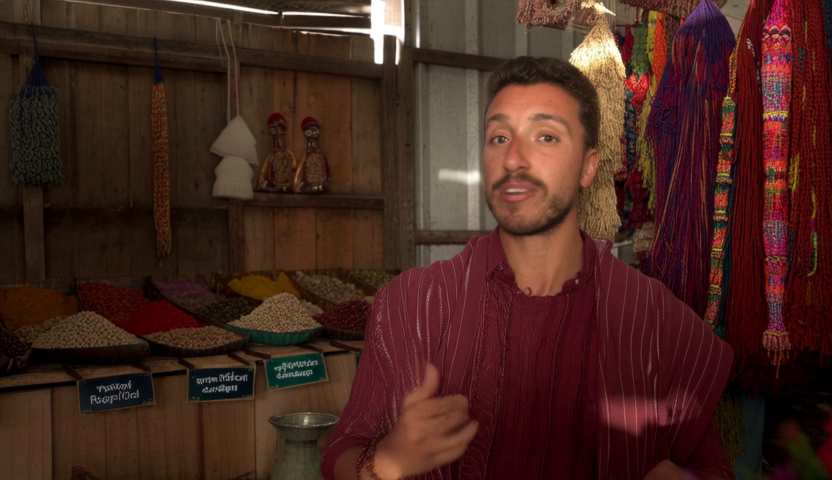}\hspace{2pt}%
  \includegraphics[width=0.32\linewidth]{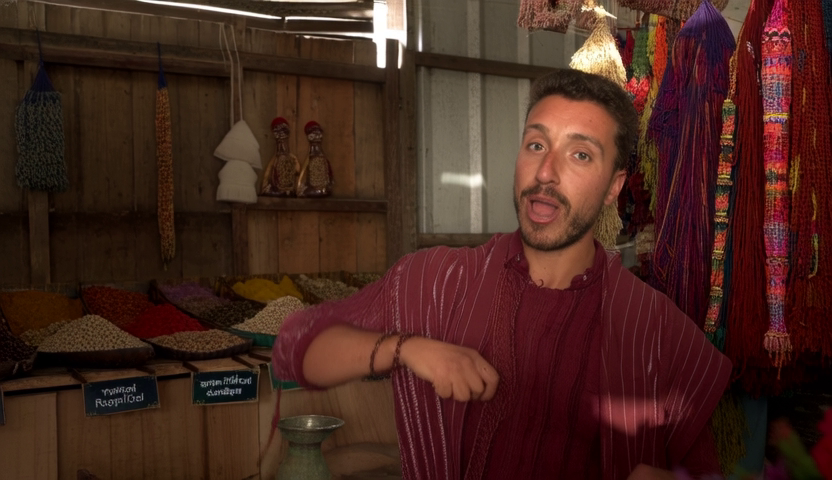}
\end{minipage}
&
&
\begin{minipage}[c]{0.43\textwidth}
  \noindent 
  \includegraphics[width=0.32\linewidth]{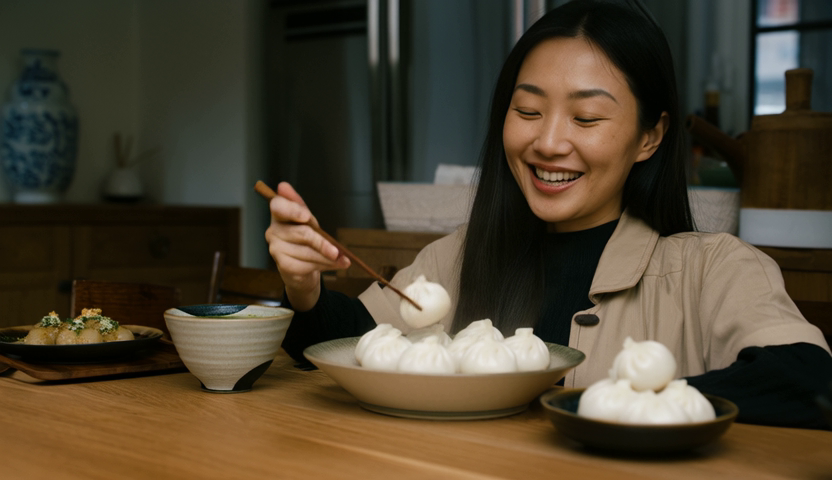}\hspace{2pt}%
  \includegraphics[width=0.32\linewidth]{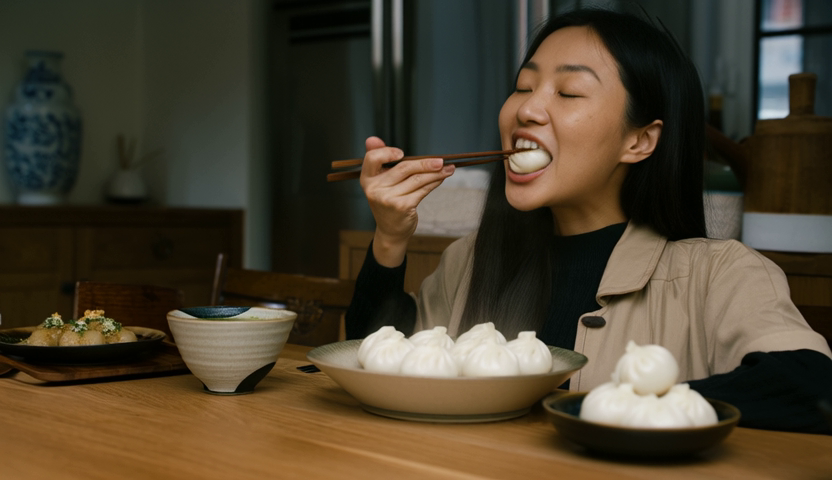}\hspace{2pt}%
  \includegraphics[width=0.32\linewidth]{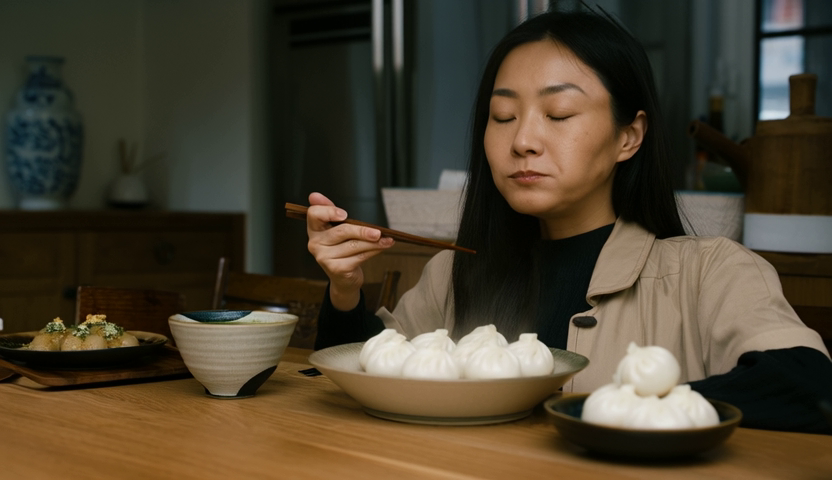}
\end{minipage}
\\[-8pt]

\parbox[c][0.12\textwidth][c]{0.06\textwidth}{\centering\rotatebox{90}{\scriptsize\bfseries Ours}}
&
\begin{minipage}[c]{0.43\textwidth}
  \noindent
  \includegraphics[width=0.32\linewidth]{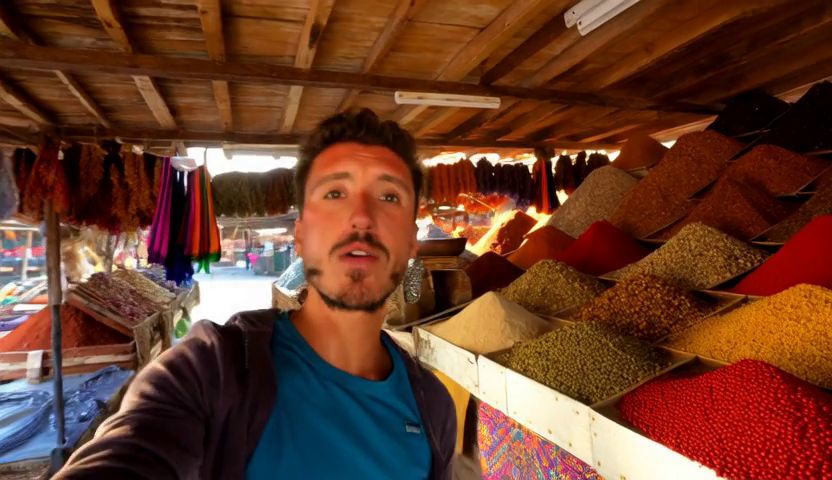}\hspace{2pt}%
  \includegraphics[width=0.32\linewidth]{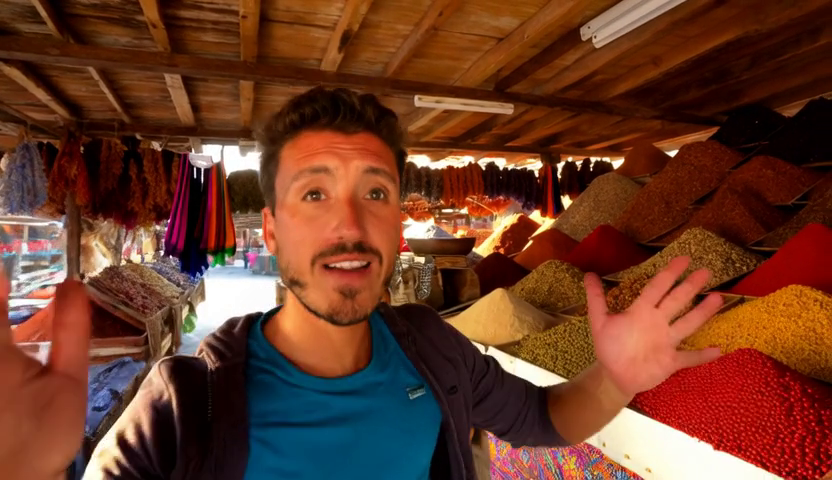}\hspace{2pt}%
  \includegraphics[width=0.32\linewidth]{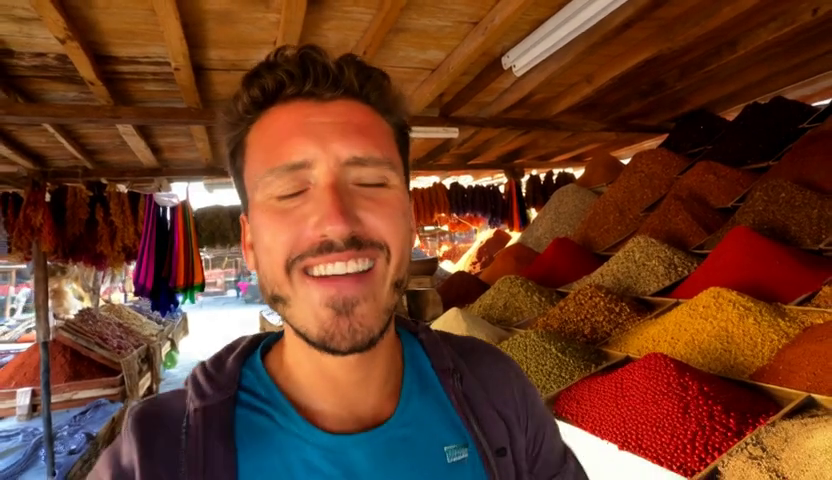}
\end{minipage}
&
&
\begin{minipage}[c]{0.43\textwidth}
  \noindent
  \includegraphics[width=0.32\linewidth]{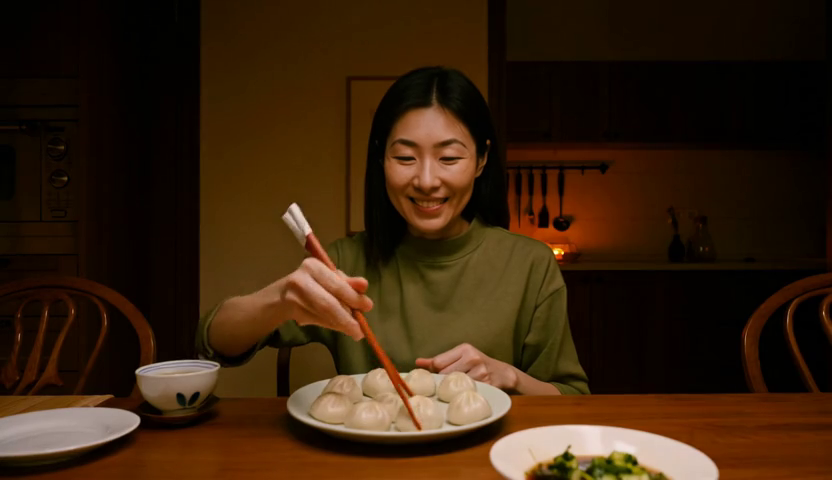}\hspace{2pt}%
  \includegraphics[width=0.32\linewidth]{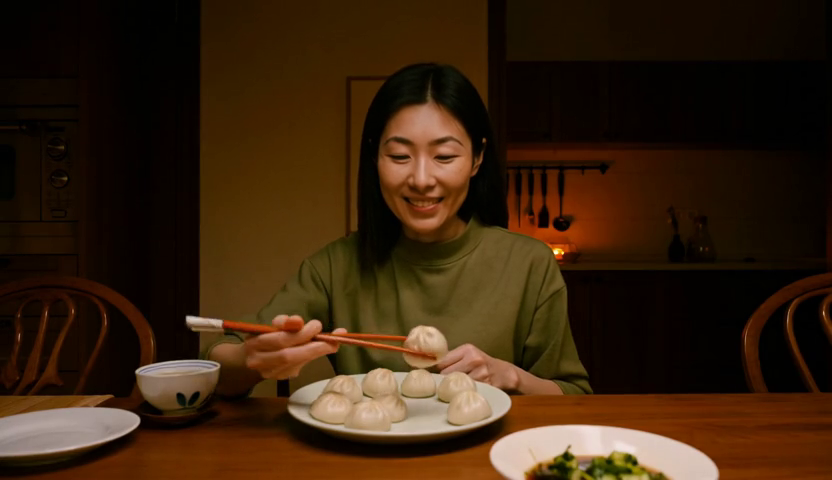}\hspace{2pt}%
  \includegraphics[width=0.32\linewidth]{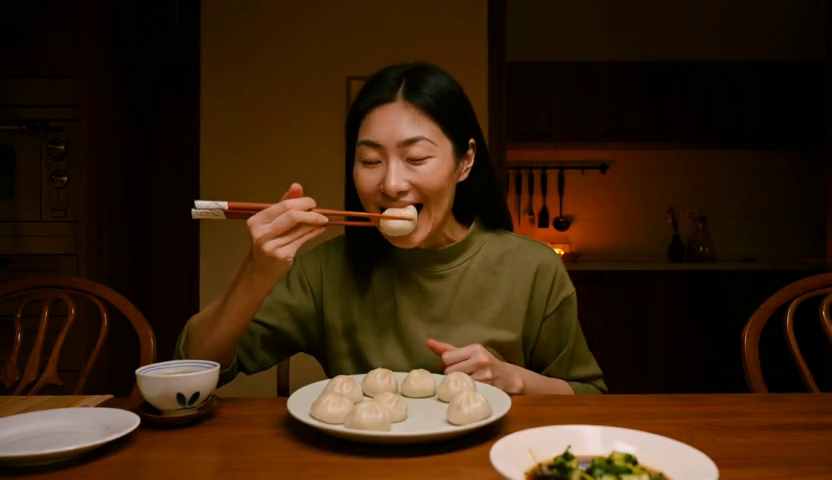}
\end{minipage}
\\
\end{tabular}
}

\caption{Qualitative comparison with baseline methods. 
Competing methods often exhibit issues such as unrealistic actions (row 1 example 2), copy-pasting effects of background (row 4 example 2) or lighting (row 5 example 2), or poor identity resemblance (row 1 example 1, row 3 example 2). In contrast, Lynx consistently preserves facial identity with high fidelity, while producing natural motion, coherent lighting, and flexible scene adaptation.}
\label{fig:visual_comp}
\end{figure}

\subsection{Benchmark and Metrics}

We construct an evaluation benchmark comprising 40 subjects and 20 unbiased text prompts, resulting in a total of 800 test videos. The subject set consists of (1) 10 celebrity photos, (2) 10 AI-synthesized portraits, and (3) 20 in-house licensed photos spanning diverse demographic groups to capture racial and ethnic diversity. The text prompts are generated using ChatGPT-4o, guided by carefully designed in-context examples, and explicitly crafted to avoid bias with respect to race, age, gender, motion, and other attributes.

We evaluate Lynx along three key dimensions: \textit{face resemblance}, \textit{prompt following}, and \textit{video quality}.  

\textbf{Face resemblance.} To measure identity fidelity, we compute cosine similarity using three independent feature extractors. These include two publicly available ArcFace implementations, facexlib\footnote{\url{https://github.com/xinntao/facexlib}} and insightface\footnote{\url{https://github.com/deepinsight/insightface}}, together with our in-house face recognition model. Employing multiple extractors reduces reliance on a single feature space and yields a more reliable assessment of identity preservation.

\textbf{Prompt following and video quality.} To assess semantic alignment and perceptual quality, we construct an automated evaluation pipeline based on the Gemini-2.5-Pro API. In this pipeline, Gemini is instructed with task-specific prompts to assign scores across four dimensions: (1) \textit{prompt alignment}, which evaluates consistency between the generated video and the input text description, (2) \textit{aesthetic quality}, which measures visual appeal and composition, (3) \textit{motion naturalness}, which captures the smoothness and realism of temporal dynamics, and (4) \textit{general video quality}, which provides an overall judgment that integrates multiple aspects of perceptual fidelity. This evaluation framework allows scalable and multi-faceted assessment of generated videos beyond traditional expert-model-based metrics.

\subsection{Qualitative Results}

Figure~\ref{fig:visual_comp} presents qualitative comparisons between Lynx and state-of-the-art baselines. As shown, existing methods frequently struggle with identity preservation, producing faces that drift away from the reference subject  or lose fine-grained details (row 1 example 1, row 3 example 2).  Moreover, they often generate unrealistic behaviors (row 1 example 2), copy-pasting effects of background (row 4 example 2) or lighting (row 5 example 2). In contrast, Lynx successfully maintains strong identity consistency across diverse prompts, while achieving natural motion, coherent visual details, and high-quality scene integration. These results demonstrate that our model effectively balances identity preservation, prompt alignment, and video realism, outperforming existing approaches both in terms of fidelity and controllability.

\subsection{Quantitative Results}

\begin{table}[h]
\centering
\caption{Quantitative comparison of Lynx with recent personalized video generation models on \textit{face resemblance}. Scores are computed with three independent evaluators: facexlib, insightface, and our in-house face recognition model. Lynx achieves the best overall identity consistency across all evaluators, while SkyReels-A2 ranks second but shows weak prompt following due to reliance on copy–paste mechanisms, as shown in Table~\ref{tab:quant_comp_gemini}.}
\label{tab:quant_comp}
\resizebox{0.5\linewidth}{!}{
\begin{tabular}{lccc}
\toprule
\multirow{2}{*}{Model} & \multicolumn{3}{c}{Face Resemblance} \\
\cmidrule(lr){2-4} & facexlib & insightface & in-house \\
\midrule
SkyReels-A2~\cite{fei2025skyreels} & \underline{0.715} & \underline{0.678} & \underline{0.725} \\
VACE~\cite{vace} & 0.594 & 0.548 & 0.615 \\
Phantom~\cite{liu2025phantom} & 0.664 & 0.659 & 0.689 \\
MAGREF~\cite{deng2025magref} & 0.575 & 0.510 & 0.591 \\
Stand-In~\cite{xue2025standin} & 0.611 & 0.576 & 0.634 \\
\midrule
Lynx (ours) & \textbf{0.779} & \textbf{0.699} & \textbf{0.781} \\
\bottomrule
\end{tabular}}
\end{table}

\begin{table}[h]
\centering
\caption{Quantitative comparison of Lynx with competing methods on \textit{prompt following}, \textit{aesthetic quality}, \textit{motion naturalness}, and \textit{overall video quality}, evaluated using the Gemini-2.5-Pro pipeline. Lynx achieves the highest performance in three out of four metrics, with particularly strong results in prompt alignment and overall quality.}
\label{tab:quant_comp_gemini}
\resizebox{0.8\linewidth}{!}{
\begin{tabular}{lcccc}
\toprule
Model & Prompt Following & Aesthetic & Motion Naturalness & Video Quality\\
\midrule
SkyReels-A2~\cite{fei2025skyreels}  & 0.471 & 0.704 & 0.824 & 0.870 \\
VACE~\cite{vace}                    & \underline{0.691} & \underline{0.846} & \textbf{0.851} & \underline{0.935} \\
Phantom~\cite{liu2025phantom}       & 0.690 & 0.825 & 0.828 & 0.888 \\
MAGREF~\cite{deng2025magref}        & 0.612 & 0.787 & 0.812 & 0.886 \\
Stand-In~\cite{xue2025standin}      & 0.582 & 0.807 & 0.823 & 0.926 \\
\midrule
Lynx (ours)                         & \textbf{0.722} & \textbf{0.871} & \underline{0.837} & \textbf{0.956} \\
\bottomrule
\end{tabular}}
\end{table}


Table~\ref{tab:quant_comp} reports quantitative comparisons across \textit{face resemblance}, \textit{prompt following}, and \textit{video quality}. On identity preservation, Lynx consistently outperforms all baselines, achieving the highest resemblance scores under facexlib, insightface, and our in-house face recognition model. SkyReels-A2 ranks second on identity resemblance, but its reliance on copy–paste generation introduces visual artifacts and leads to weak semantic alignment, as reflected in its poor prompt following performance as shown in Table~\ref{tab:quant_comp_gemini}. Phantom demonstrates strong prompt alignment but does so at the expense of identity fidelity, suggesting a trade-off between semantic consistency and subject preservation. In contrast, Lynx achieves the best balance, combining superior identity fidelity with competitive prompt alignment, highlighting the advantage of our adapter-based design.  

Table~\ref{tab:quant_comp_gemini} further evaluates \textit{prompt following}, \textit{aesthetic quality}, \textit{motion naturalness}, and \textit{overall video quality} using the Gemini-2.5-Pro evaluation pipeline. Lynx delivers the best performance in four out of five metrics, including prompt alignment, aesthetics, and overall video quality, which demonstrates the perceptual quality of our outputs. VACE attains the highest score in motion naturalness, reflecting its strong temporal modeling capability, while Phantom and Stand-In perform competitively across most dimensions but lag behind in overall video quality. These results show that Lynx not only preserves identity more effectively but also produces videos that are semantically accurate, visually appealing, and of high perceptual quality.

Figure~\ref{fig:radar_chart} provides a visual summary of these comparisons, where Lynx demonstrates consistent superiority across identity resemblance and perceptual quality dimensions, while remaining competitive in motion naturalness. The combined evidence from multiple evaluators underscores the robustness of our approach and establishes Lynx as a new state of the art in personalized video generation.
\section{Conclusion}
\label{sec:conclusion}

In this work, we introduced Lynx, a high-fidelity framework for personalized video generation that preserves subject identity from a single reference image. The model incorporates two lightweight adapters: the ID-adapter, which encodes ArcFace-derived identity tokens, and the Ref-adapter, which integrates VAE-based dense features through a frozen reference pathway. Together, these components enable robust identity fidelity while maintaining motion naturalness and visual coherence. We evaluated Lynx on a curated benchmark of 40 subjects and 20 unbiased prompts, totaling 800 test cases across diverse identities and scenarios, and found that it achieves state-of-the-art performance in face resemblance while also delivering competitive prompt following and strong video quality. Overall, Lynx advances personalized video generation with a scalable adapter-based framework that balances identity preservation, controllability, and realism, and lays the groundwork for future extensions toward multi-modal and multi-subject personalization.

\clearpage

\bibliographystyle{plainnat}
\bibliography{main}

\end{document}